%% file: main.tex
\documentclass{article}

\PassOptionsToPackage{sort, numbers, compress}{natbib}

\usepackage[preprint]{neurips_2026}

\usepackage[utf8]{inputenc}
\usepackage[T1]{fontenc}
\usepackage{hyperref}
\usepackage{url}
\usepackage{booktabs}
\usepackage{array}
\usepackage{longtable}
\usepackage{amsfonts}
\usepackage{nicefrac}
\usepackage{microtype}
\usepackage[table]{xcolor}
\usepackage{tcolorbox}
\tcbuselibrary{breakable}

\usepackage{amsmath}
\usepackage{pifont}
\usepackage{graphicx}
\usepackage{multirow}
\usepackage{makecell}
\usepackage{enumitem}
\usepackage{placeins}
\newcommand{\benchname}{WordEval}
\newcommand{\wordobj}{Word Object}
\newcommand{\gtdoc}{reference document}
\input{assets/stats}

\title{
WordEval: Evaluating Word-Native Operation Fidelity in Document Editing
}

\author{%
  Jinyoung~Park\textsuperscript{1\textdagger*}\quad
  Yupan~Huang\textsuperscript{2\textdagger}\quad
  Shimin~Zhang\textsuperscript{2}\quad
  Jiayu~Ding\textsuperscript{3*}\quad
  Yilin~Jia\textsuperscript{2}\\
  \bfseries Yuzhong~Zhao\textsuperscript{2}\quad
  Tengchao~Lv\textsuperscript{2}\quad
  Wenshan~Wu\textsuperscript{2}\quad
  Shaohan~Huang\textsuperscript{2}\quad
  Nan~Yang\textsuperscript{2}\\
  \bfseries Xiangyang~Zhou\textsuperscript{2}\quad
  Dongdong~Zhang\textsuperscript{2}\quad
  Li~Dong\textsuperscript{2}\quad
  Lei~Cui\textsuperscript{2}\quad
  Sun~Mao\textsuperscript{2}\\
  \bfseries Xun~Wang\textsuperscript{2}\quad
  Tao~Ge\textsuperscript{2}\quad
  Huitian~Jiao\textsuperscript{2}\quad
  Si-Qing~Chen\textsuperscript{2}\quad
  Changick~Kim\textsuperscript{1\textdaggerdbl}\quad
  Furu~Wei\textsuperscript{2\textdaggerdbl}\\
  \textnormal{\textsuperscript{1}KAIST\quad \textsuperscript{2}Microsoft\quad \textsuperscript{3}XJTU}\\[0.3em]
  \textnormal{\small\url{https://aka.ms/word-eval}}
}

\makeatletter
\renewcommand{\@thanks}{%
  \footnotetext[2]{Equal contribution.\quad\textsuperscript{\textdaggerdbl}Corresponding authors.}%
  \footnotetext[1]{Work done during an internship at Microsoft Research.}%
}
\makeatother

\begin{document}
\maketitle
\addtocontents{toc}{\protect\setcounter{tocdepth}{-10}}

\input{sections/00_abstract}
\input{sections/01_introduction}
\input{sections/02_rel_works}

\input{sections/03_method}
\input{sections/04_experiments}
\input{sections/05_conclusion}

\renewcommand{\acksection}{\section*{Acknowledgments}}
\begin{ack}
This research was supported by the MSIT (Ministry of Science, ICT), Korea, under the Global Research Support Program in the Digital Field program (RS-2024-00436680) supervised by the IITP (Institute for Information \& Communications Technology Planning \& Evaluation).
This project is supported by Microsoft Research Asia.
\end{ack}
\clearpage

\input{reference.bbl}
\clearpage
\appendix
\addtocontents{toc}{\protect\setcounter{tocdepth}{2}}
\input{sections/99_appendix}

\end{document}

%% file: assets/stats.tex
\newcommand{\AtomicCompositionGapMax}{19.1}
\newcommand{\AtomicCompositionGapMin}{3.2}
\newcommand{\EvalOSBuild}{10.0.26200}
\newcommand{\EvalPythonVersion}{3.12}
\newcommand{\EvalWordBuild}{16.0.17932.21000}
\newcommand{\FeaturePageStructureMean}{53.8}
\newcommand{\FeatureParagraphsMean}{81.1}
\newcommand{\FeaturePicturesMean}{44.9}
\newcommand{\FeatureSmartArtMean}{50.7}
\newcommand{\FeatureTablesMean}{83.2}
\newcommand{\GLMAtomicCase}{58.2}
\newcommand{\GLMCompositionCase}{39.6}
\newcommand{\GLMLayoutConditionalCase}{25.0}
\newcommand{\GPTAdditiveConditionalGain}{2.3}
\newcommand{\GPTInvalidMissingShare}{18.4}
\newcommand{\GPTLayoutCase}{76.9}
\newcommand{\GPTLayoutConditionalCase}{66.7}
\newcommand{\GeminiAtomicCase}{57.9}
\newcommand{\GeminiCompositionCase}{49.0}
\newcommand{\GeminiMissingShare}{41.7}
\newcommand{\HumanOverallCase}{82.8}
\newcommand{\KimiAdditiveConditionalGain}{41.6}
\newcommand{\OpusAtomicCase}{87.5}
\newcommand{\OpusCompositionCase}{84.4}
\newcommand{\OpusContentShare}{6.8}
\newcommand{\OpusMissingElementShare}{11.9}
\newcommand{\OpusOffTargetShare}{6.8}
\newcommand{\OpusOverallCase}{87.1}
\newcommand{\OpusPropertyValueShare}{39.0}
\newcommand{\OpusStructureShare}{33.9}
\newcommand{\OpusUnrequestedShare}{1.7}
\newcommand{\OtherAgentOverallCaseMax}{66.7}
\newcommand{\OtherAgentOverallCaseMin}{44.5}
\newcommand{\SelectedLayoutCaseMax}{84.6}
\newcommand{\SelectedLayoutCaseMin}{23.1}
\newcommand{\SelectedScopeCaseMax}{100}
\newcommand{\SelectedScopeCaseMin}{58.3}
\newcommand{\nAdditiveCases}{46}
\newcommand{\nAtomicCases}{546}
\newcommand{\nAtomicCommands}{231}
\newcommand{\nCases}{642}
\newcommand{\nCompCases}{96}
\newcommand{\nCompTemplates}{27}
\newcommand{\nInScopeCommands}{992}
\newcommand{\nIntentDeclared}{2552}
\newcommand{\nIntentPairs}{1701}
\newcommand{\nLayoutCoupledCases}{13}
\newcommand{\nNormalizedCommands}{2{,}569}
\newcommand{\nOpusIncorrectOutputs}{59}
\newcommand{\nOpusNonPassing}{83}
\newcommand{\nPrerequisiteCases}{25}
\newcommand{\nRibbonControlEntries}{5{,}638}
\newcommand{\nScopeChangingCases}{12}
\newcommand{\nSubsystems}{15}
\newcommand{\nVerifiableCommands}{1{,}150}

%% file: sections/00_abstract.tex
\begin{abstract}
Microsoft Word is a central environment for office automation, and the faithful execution of its editing operations is an important capability for document agents.
Unlike plain-text editing, editing in Word modifies a structured object model, where the requested effect is realized in object properties.
Existing benchmarks, however, evaluate office workflows or desktop automation, treating word-processor documents as one task domain among many and verifying outcomes with keyword matching, per-task property checks, or screenshot comparison.
We introduce \textbf{\texttt{\benchname{}}}, an operation-fidelity benchmark that instead centers evaluation on Word editing operations, testing whether each requested edit is realized as intended changes to Word objects and properties.
It contains \nCases{} cases across \nSubsystems{} Word feature areas, ranging from text and lists to charts, shape geometry, and SmartArt.
\nAtomicCases{} atomic cases cover \nAtomicCommands{} commands, \nCompCases{} compositional cases combine multiple edits in a single document, and a Word-native oracle checks task-specific object properties and preservation conditions.
Cases are constructed from editing-unit specifications, yielding controlled input documents, reference documents, and aligned oracle checks.
Each compositional case is linked to corresponding atomic cases, supporting fine-grained diagnosis of editing failures.
We evaluate eight agent systems and find that many struggle on the Atomic track, with performance varying across feature areas.
Moreover, every system scores lower on Compositional than on Atomic, with gaps that vary substantially by system. Exact atomic-intent links and recorded failure diagnoses provide complementary views of this variation.
\end{abstract}

%% file: sections/01_introduction.tex
\begin{figure}[t]
\centering
\includegraphics[trim=0 15.16bp 0 22.74bp,clip,width=\linewidth]{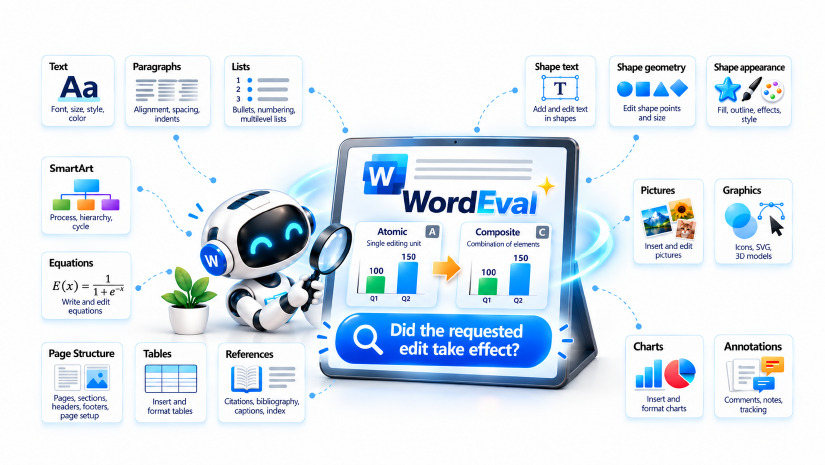}
\caption{Overview of \benchname{}. Atomic tasks test individual Word-editing operations; Compositional tasks combine linked operations across \nSubsystems{} feature areas. The evaluator checks the resulting document's Word-native objects and properties against the reference document to determine whether the requested edits were realized.}
\label{fig:wordeval-overview}
\end{figure}

\section{Introduction}

AI assistants have begun to operate inside the software where users work, including word processors such as Microsoft Word~\cite{claudecode,codex,githubcopilot}. 
In this context, their output is no longer merely a natural-language reply, but a document that the user or agent itself can create, edit, and reuse. 
A Word document, however, is not merely a text canvas: it contains underlying information that the word processor reads and applies across subsequent document operations. 
A heading can be used for navigation or table-of-contents generation, a list can maintain continuous numbering as items are added or removed, and a field can be dynamically updated. 
Therefore, the central question is whether the requested operation takes effect in the document in a native form that Word can recognize and apply.

Current evaluations of document agents, however, do not isolate this ability well.
Existing benchmarks largely evaluate document generation, file I/O, information extraction, or GUI task completion~\cite{officebench,odysseybench,osworld,wagentarena,worldgui}.
They show that an agent can access a document and produce an output file, but not whether the effect of each Word-editing operation was faithfully realized.
To address this gap, we introduce \benchname{}, a benchmark that evaluates how faithfully a Word agent realizes a user's editing intent as a change in document state.
We call this ability \emph{operation fidelity}: whether the intended effect of an editing operation is realized in the resulting document as a state change with the correct target, attribute, and parameters.
\benchname{} evaluates the output document itself, independent of the action path (e.g., GUI clicks, ribbon commands, or API calls) used to produce it.

\benchname{} organizes evaluation into Atomic and Compositional tracks.
Atomic cases isolate whether the intended effect of a single editing operation is realized. 
Compositional cases examine whether the agent can manage dependencies, ordering, interference, and preservation when multiple edits are issued together.
This design lets us separately analyze the ability to perform individual Word-editing operations and to integrate multiple edits into a coherent document state.

\paragraph{Contributions.}
\textbf{(i) A Word-native operation-fidelity benchmark.} \benchname{} comprises \nCases{} cases spanning \nSubsystems{} Word feature areas, each generated from an editing-unit specification with an aligned Word-native oracle, with an Atomic track of \nAtomicCases{} cases over \nAtomicCommands{} Word commands and a Compositional track of \nCompCases{} cases from \nCompTemplates{} composition scenarios.

\textbf{(ii) Evaluation revealing a compositional gap.} Across eight agent systems, many still struggle on the Atomic track, with system-specific Atomic-to-Compositional gaps.

\textbf{(iii) Fine-grained failure diagnosis.} Atomic-intent links, which connect each editing unit of a Compositional case to the Atomic case with the same command, trace compositional failures to individual operations, and a failure-mode analysis characterizes how incorrect outputs differ from the reference.

\begin{figure}[t]
\centering
\includegraphics[width=\linewidth]{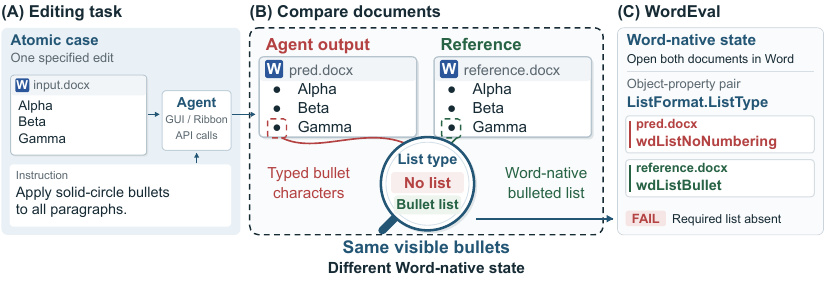}
\caption{An illustrative example of Word-native operation-fidelity evaluation. In this example, the document agent inserts bullet characters without creating a list in Word, yielding an output visually similar to the reference. \benchname{} detects this property mismatch.}
\label{fig:framework}
\end{figure}

%% file: sections/02_rel_works.tex
\section{Related Work}
\textbf{Document editing in office benchmarks.}
Recent benchmarks for office-workflow agents span desktop and office settings.
OfficeBench~\citep{officebench} and OdysseyBench~\citep{odysseybench} evaluate multi-application office workflows through file-level I/O and conversion utilities, while GUI benchmarks such as OSWorld~\citep{osworld}, WindowsAgentArena~\citep{wagentarena}, and WorldGUI~\citep{worldgui} evaluate desktop automation with word processors such as Microsoft Word or LibreOffice Writer appearing as one target application.
These benchmarks show that agents can reach and manipulate document files, but they do not provide a controlled measurement of document-editing capability organized around the space of editing operations.
\benchname{} targets this missing layer by evaluating structured edits in Microsoft Word under controlled specifications drawn from a command-level taxonomy.

\textbf{Verification in office and GUI benchmarks.}
Existing protocols typically verify returned text, file contents, saved-file state, task completion, or visual appearance: OfficeBench~\citep{officebench} and OdysseyBench~\citep{odysseybench} use exact-match, containment, keyword, and file-existence checks; OSWorld~\citep{osworld} and WindowsAgentArena~\citep{wagentarena} apply plain-text matching and per-property checks to saved files; and WorldGUI~\citep{worldgui} evaluates screenshot-level GUI task success.
These evaluators are effective for checking task completion, but they do not consistently verify the target Word-native object state across commands.
For example, a numbered list can be approximated by typing plain-text markers such as \texttt{1.} and \texttt{2.}, matching the rendered appearance without behaving as a Word list under subsequent renumbering or level changes.
\benchname{} instead evaluates the resolved document state, checking that the specified change holds on the target object and that surrounding content and structure are preserved.
Table~\ref{tab:positioning} summarizes this comparison.

\input{assets/tables/benchmark_comparison}

%% file: assets/tables/benchmark_comparison.tex
\begin{table}[!t]
\centering
\caption{Comparison with prior benchmarks containing word-processing tasks. Interface is the route through which an agent edits the document. \benchname{} agents choose their own route, such as \texttt{python-docx}, VBA, or the Word GUI. $^{*}$Estimated from Word-document subtasks. $^{\dagger}$Variants of 11 meta-tasks.}
\label{tab:positioning}
\setlength{\tabcolsep}{3.0pt}
\renewcommand{\arraystretch}{1.15}
\footnotesize
\begin{tabular}{l l r l l}
\toprule
\textbf{Benchmark} & \textbf{Interface} & \textbf{Word tasks} & \textbf{Edit operations} & \textbf{Verification oracle} \\
\midrule
OfficeBench~\citep{officebench} & \texttt{python-docx} API & 80\rlap{$^{*}$} & 4 file actions & keyword, file existence \\
OdysseyBench~\citep{odysseybench} & \texttt{python-docx} API & 235\rlap{$^{*}$} & 4 file actions~\citep{officebench} & keyword, file existence \\
OSWorld~\citep{osworld} & LibreOffice GUI & 23 & $\approx$23 property edits & text match, property checks \\
WindowsAgentArena~\citep{wagentarena} & LibreOffice GUI & 19 & 18 property edits~\citep{osworld} & text match, property checks \\
WorldGUI~\citep{worldgui} & Word GUI & 63\rlap{$^{\dagger}$} & 11 meta-tasks & screenshot task success \\
\midrule
\rowcolor{gray!10}
\textbf{\benchname{} (ours)} & Unrestricted & \nCases{} & \nAtomicCommands{} commands & Word object properties, preservation \\
\bottomrule
\end{tabular}
\end{table}

%% file: sections/03_method.tex
\section{\benchname{}}

\subsection{Problem Formulation}
\label{sec:task_formulation}

\paragraph{The Word object model.}
Word represents a document as a structured object graph, which we call the \emph{Word object model}.
In this model, Word exposes runtime objects such as \texttt{Shape}, \texttt{Style}, \texttt{Font}, \texttt{Paragraph}, \texttt{Field}, and \texttt{Comment}, together with properties including \texttt{Shape.Rotation}, \texttt{Style.NameLocal}, and \texttt{Font.Bold}.
These runtime objects and their property values are resolved when Word opens and interprets a document.

\paragraph{Operation fidelity.}
\benchname{} evaluates \emph{Word-native operation fidelity}, an agent's ability to realize a requested edit as the intended changes to objects and properties in the Word object model.
We formalize each evaluation case as a tuple $(d_0,u,d^*)$, where $d_0$ is the input document, $u$ is a natural-language instruction, and $d^*$ is a reference document representing the expected post-edit state.
Given $(d_0,u)$, the agent produces an edited document $\hat{d}$.
We evaluate $\hat{d}$ using the Word-native oracle described in Section~\ref{sec:eval}, which verifies the required target changes and preservation properties.
This evaluation is independent of the action sequence and specific Word commands used to produce $\hat{d}$.

\begin{table}[!t]
    \centering
    \caption{Structure and coverage of \benchname{}, including Atomic and Compositional case counts, Word feature areas, and composition classes.}
    \label{tab:benchmark-structure}
    \scriptsize
    \setlength{\tabcolsep}{3.5pt}
    \renewcommand{\arraystretch}{1.05}
    \begin{tabular}{@{}>{\raggedright\arraybackslash}p{0.165\linewidth}r>{\raggedright\arraybackslash}p{0.355\linewidth}>{\raggedright\arraybackslash}p{0.335\linewidth}@{}}
        \toprule
        \textbf{Category} & \textbf{Cases} & \textbf{Coverage} & \textbf{Definition} \\
        \midrule
        \textbf{\benchname{}} & \nCases{} & \nAtomicCommands{} commands; \nCompTemplates{} scenarios & Benchmark across \nSubsystems{} Word feature areas with task-specific object-state checks. \\
        \midrule
        \textbf{Atomic} & \nAtomicCases{} & \nAtomicCommands{} commands & Single editing unit over one Word command. \\
        \hspace*{0.7em}Feature areas & \nAtomicCases{} & \nSubsystems{} feature areas (\texttt{Text}, \texttt{Paragraphs}, \texttt{Lists}, \texttt{Tables}, \texttt{Shape geometry}, \texttt{Shape appearance}, \texttt{Shape text}, \texttt{Pictures}, \texttt{Graphics}, \texttt{Charts}, \texttt{SmartArt}, \texttt{Equations}, \texttt{Page Structure}, \texttt{Annotations}, \texttt{References}) & Object-model coverage axis; feature areas group runtime objects, which expose checked object-property pairs. \\
        \midrule
        \textbf{Compositional} & \nCompCases{} & \nCompTemplates{} scenarios & Multiple editing units in the same controlled document substrate. \\
        \hspace*{0.7em}Additive & \nAdditiveCases{} & 11 scenarios: font, paragraph, list, document enrichment/cleanup, chart, layout, picture, shape, SmartArt, table & Unit preconditions are jointly satisfied and edits do not invalidate one another. \\
        \hspace*{0.7em}Interaction-sensitive & 50 & 16 scenarios & Earlier edits affect later preconditions, target scope, or layout state. \\
        \hspace*{1.1em}Prerequisite-chain & \nPrerequisiteCases{} & 7 scenarios: tables, charts, icons, shapes, SmartArt, equations, text boxes & Later unit requires an object or state created by an earlier unit. \\
        \hspace*{1.1em}Scope-changing & \nScopeChangingCases{} & 4 scenarios: tables, SmartArt, equations, lists & Earlier unit changes the target set or structural scope used by later units. \\
        \hspace*{1.1em}Layout-coupled & \nLayoutCoupledCases{} & 5 scenarios: grouped shapes, shape rotation/alignment, shape position/alignment, shape multi-step geometry, text wrapping/positioning & Later geometry or layout checks depend on earlier wrapping, alignment, or positioning state. \\
        \bottomrule
    \end{tabular}
\end{table}
\FloatBarrier

\subsection{Benchmark Structure and Coverage}
\label{sec:operation_space}
\label{sec:capability}

\paragraph{Editing unit as case specification.}
We represent each edit as an \emph{editing unit} defined by four components: a precondition that makes the edit executable, a target span or runtime object, an operation, and the parameter values for that operation.
The precondition specifies the document state required for the operation to have a determinate, verifiable effect, such as an existing target object, an applicable selection, or a structural context.
The operation is represented by a normalized Word command, whereas the target identifies the span or runtime object to which that command applies.

\paragraph{Operation taxonomy.}
The \textit{operation taxonomy} organizes the normalized Word commands populating the operation component of each editing unit, associating each command with a generation route and verification schema.
Built by consolidating redundant Word ribbon metadata\footnote{\href{https://github.com/OfficeDev/office-fluent-ui-command-identifiers/blob/main/Microsoft\%20365/Current\%20Channel/wordcontrols.xlsx}{OfficeDev wordcontrols.xlsx}} into distinct semantic commands (Appendix~\ref{app:taxonomy}), each entry follows a three-tier hierarchy: \textit{behavior family} (generation logic), \textit{target Word component} (affected state), and \textit{verification schema} (object-property oracle comparisons).
This structure enables systematic case generation with aligned oracle specifications (Section~\ref{sec:eval}).
We exclude commands that are purely UI-structural, service-dependent, lack deterministic and verifiable effects, or fall outside our scope.
Among the remaining commands, we instantiate those whose preconditions can be constructed, operations automated, and postconditions verified.
This yields \nAtomicCommands{} released commands; Appendix~\ref{app:outofcoverage} details the coverage funnel and exclusion criteria.

\paragraph{Word feature areas.}
Whereas the operation taxonomy organizes what edit is performed, Word feature areas organize where that edit acts in the Word object graph.
We assign each command to a feature area by its target runtime object.
The released commands span \nSubsystems{} Word feature areas, from \texttt{Text}, \texttt{Paragraphs}, and \texttt{Tables} to \texttt{Charts}, \texttt{SmartArt}, and \texttt{Page Structure} (Table~\ref{tab:benchmark-structure}).

\paragraph{Evaluation tracks.}
To distinguish fidelity on individual edits from the additional burden of composition, we organize cases into two tracks, \emph{Atomic} and \emph{Compositional}.
The Atomic track contains cases with a single editing unit, whereas compositional cases combine multiple units within the same controlled document substrate.
Both tracks use the same operation pool and the same Word-native evaluation criterion, allowing individual and combined edits to be compared under one standard.
We group compositional cases into two categories by how their editing units interact, \emph{additive} and \emph{interaction-sensitive}.
In additive cases, unit preconditions are jointly satisfied in $d_0$, no unit invalidates another, and the units are order-independent.
In interaction-sensitive cases, earlier edits affect later ones, and three classes specify the affected state, namely preconditions in \emph{prerequisite-chain} cases, target scope in \emph{scope-changing} cases, and layout state in \emph{layout-coupled} cases.
The additive category and these three classes form the four composition classes used in our analyses.
Appendix Tables~\ref{tab:atomic-case-examples} and~\ref{tab:compositional-case-examples} list representative released instructions.
For example, an atomic case asks the agent to make the word ``Target'' bold, while a prerequisite-chain case asks it to insert a clustered column chart below the ``Quarterly Sales:'' heading and then add outside-end data labels.

Table~\ref{tab:benchmark-structure} summarizes the resulting benchmark structure, case counts, and coverage over Word feature areas and composition classes.
Appendix~\ref{app:object_coverage} gives the object-property coverage, and \S\ref{sec:capability_profile} reports capability stratified by feature area.

\begin{figure}[!t]
    \centering
    \includegraphics[width=\linewidth]
    {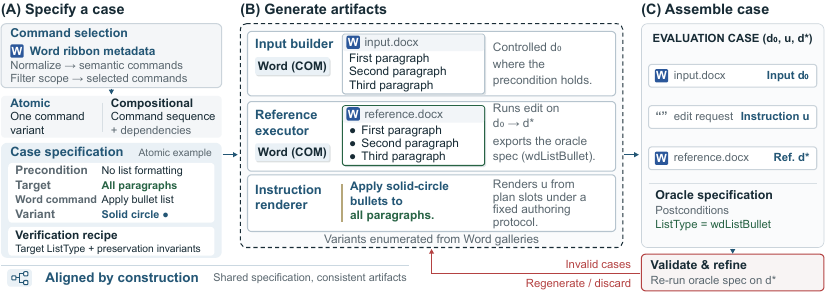}
    \caption{Case generation pipeline. A shared case source specifies the editing unit or unit sequence, preconditions, target objects, parameters, and verification schema. The generator uses this source to materialize the input document, instruction, \gtdoc{}, metadata, and oracle specification.}
    \label{fig:benchmark-construction}
\end{figure}

\subsection{Case Construction Pipeline}
\label{sec:generation}

\paragraph{Generation pipeline.}
Figure~\ref{fig:benchmark-construction} shows the end-to-end pipeline.
Given a planned editing unit and its schema, an \emph{Input Builder} constructs a controlled input document $d_0$ in which the unit's precondition holds, materializing the required document state explicitly rather than sampling it from natural documents so that the input contains only the structure required by the operation under test.
A \emph{Reference Executor} then runs the planned operation on $d_0$ under Word's native automation, dispatching to the invocation route appropriate for the command (one-shot ribbon execution, parameter selection, or dialog automation), producing the reference document $d^*$, and exporting the oracle specification defined by the schema.
An oracle specification lists the checks applied to one case, covering target postconditions for the requested edit and preservation checks for state the edit should leave unchanged.
In parallel, an \emph{Instruction Renderer} produces the natural-language instruction $u$ from the same unit's targets, operation, and parameters under a fixed authoring protocol: each unit names a single operation, references a concrete target visible in $d_0$, and avoids transient or invisible state such as selection or cursor focus.
Together these stages yield a complete case $(d_0, u, d^*)$ with its oracle specification attached.
The generator also enumerates parameter variants through Word COM automation, so scores are not tied to a single parameter setting; Appendix~\ref{app:construction_details} details the variant-generation procedure.
The pipeline drives both case construction and oracle evaluation through Word's native automation, a symmetric COM-API design that makes the target postconditions executable.
Properties manipulated during reference construction can be read back during verification, which avoids inferring these postconditions from text, screenshots, or LLM judgments.

\paragraph{Compositional case construction.}
We derive compositional cases from the same document substrates and operations as atomic cases and add dependencies across edits, keeping the operations fixed while varying how they are combined.
For atomic cases, the unit's precondition is satisfied by $d_0$ alone.
For compositional cases, the precondition extends across the sequence and is distributed according to the composition class.
In additive compositions, $d_0$ is constructed so that all unit preconditions are jointly satisfied and no unit invalidates another's precondition.
In interaction-sensitive compositions, $d_0$ satisfies only the initial precondition, and later preconditions are satisfied only by states produced by earlier units, with the specific carry-over pattern determined by the interaction class (prerequisite-chain, scope-changing, or layout-coupled).

\paragraph{Atomic-intent links.}
\label{sec:atomic_intents}
An \emph{atomic intent} is one editing unit of a compositional case, linked to the Atomic case that applies the same command.
In the chart case of \S\ref{sec:operation_space}, inserting the chart and adding outside-end data labels are two atomic intents, linked to the \texttt{ChartInsert} and \texttt{ChartDataLabel} Atomic cases.
Section~\ref{sec:composition_burden} uses these links to examine compositional cases for which the same system passes every linked Atomic case.

\paragraph{Quality control.}
Before release, the oracle applies each case's specification to the reference document $d^*$ itself, and every check must pass.
All 642 reference documents pass these checks.
Cases that fail this self-check, produce inconsistent metadata, or violate the authoring protocol for $u$ are regenerated or discarded.
This step checks consistency across $d_0$, $u$, $d^*$, and the oracle; released cases are also reviewed by human annotators for instruction clarity, input validity, and reference correctness (Appendix~\ref{app:quality_control}).

\subsection{Word-Native Oracle}
\label{sec:eval}
Each case defines a set of property comparisons, implemented as our oracle.
A comparison reads a value from the agent output $\hat{d}$ and checks it against a reference value obtained from $d^*$, $d_0$, or case metadata, and a case passes only if all required comparisons pass.
The oracle reads these values through Word's object model; for some commands, it also checks the saved Office Open XML package to confirm that the requested state persists in the saved file.
The oracle verifies task-specific object properties and preservation conditions across \nSubsystems{} Word feature areas. Appendix~\ref{app:verification} details coverage, schemas, and human validation. Appendix Table~\ref{tab:capability-hierarchy} summarizes the oracle's object--property coverage across Word feature areas; each case's oracle specification defines the applicable checks.
These comparisons cover object attributes, structure counts, inserted content, reference positions, and preservation properties.
Each schema pairs its target postcondition with invariants for non-target state that the operation should leave unchanged, such as surrounding text, paragraph or object counts, section structure, table dimensions, or field counts when those properties are relevant.
This prevents narrow target-only passes while keeping preservation schema-scoped rather than a global whole-document equivalence rule.
For compositional cases, the set combines unit-level checks for individual units with case-level checks for cross-unit dependencies.
Additive compositions require no case-level checks, whereas interaction-sensitive compositions include checks specific to the interaction class.
Comparisons use behavior-specific extractors and tolerances, including numerical tolerances for position-related properties.
When a document opens in Word, each failed check reports the compared property with its reference and output values, such as a footnote number format of lowercase letters in the reference and Arabic numerals in the output.
These reports ground the failure-mode analysis in \S\ref{sec:failure_modes}.

%% file: sections/04_experiments.tex
\section{Experiments and Analysis}
\label{sec:experiments}
\subsection{Experimental Setup}
\label{sec:setup}
\input{assets/tables/main_results}
\paragraph{Evaluated systems.}
We evaluate eight agent systems: Claude Code~\citep{claudecode} with Opus~5~\citep{claudeopus5}; Codex~\citep{codex} with GPT-5.6 SOL~\citep{gpt56} (high); Kimi Code~\citep{kimicode} with K3~\citep{kimik3} (high); Mini Agent~\citep{miniagent} with MiniMax~M3~\citep{minimaxm3}; Qwen Code~\citep{qwencode} with Qwen3.8-27B~\citep{qwen3827b}; OpenCode~\citep{opencode} with Gemini~3.8 Flash~\citep{gemini38flash} (high); ZCode~\citep{zcode} with GLM-5.2~\citep{glm52} (high); and OpenCode~\citep{opencode} with DeepSeek V4 Pro~\citep{deepseekv4pro}. Each system is evaluated on 642 cases: 546 Atomic and 96 Compositional. Three human participants perform the same task set.
\paragraph{Metrics.}
We report two \wordobj{} pass-rate metrics.
The \textit{case-level} pass rate weights every case equally.
The \textit{macro} pass rate is the unweighted mean over atomic commands or composition scenarios, each scored by its variant mean.
Each command or composition scenario thus contributes equally; feature-area breakdowns report macro rates unless otherwise stated.
Missing outputs, including agent runs that end without a document, and invalid outputs count as non-passing cases (\S\ref{sec:failure_modes}).

\paragraph{Execution environment.}
Benchmark generation, reference construction, and \wordobj{} verification run on Windows with Microsoft Word through COM automation; these stages require no GPU.
The reported run used Windows~11 virtual machines with Word~16.0 and 8GB RAM.
Agent-side inference is provided by each agent system and may use external model-serving infrastructure.

\paragraph{Artifact and reproducibility.}
\benchname{} is a self-contained release whose evaluator needs no agent system; Appendix~\ref{app:artifact} details its contents, licenses, data card, and environment requirements.

\subsection{Main Results}
\label{sec:main_results}
Table~\ref{tab:main-results} reports the performance of eight agent systems on Atomic and Compositional tasks. Claude Code with Opus~5 achieves the highest Overall Case pass rate at \OpusOverallCase{}\%, while the remaining systems score between \OtherAgentOverallCaseMin{}\% and \OtherAgentOverallCaseMax{}\%. The three human participants achieve a mean Overall Case pass rate of \HumanOverallCase{}\% on the same tasks. Individual participant scores and the human execution protocol are reported in Table~\ref{tab:human-execution} and Appendix~\ref{app:agent_setup}.

Across the two tracks, all eight systems have lower Compositional than Atomic Case pass rates, but the gap varies from \AtomicCompositionGapMin{} to \AtomicCompositionGapMax{} percentage points. Opus~5 maintains high performance on both tracks, with pass rates of \OpusAtomicCase{}\% and \OpusCompositionCase{}\%, respectively. However, systems with similar performance on individual edits do not necessarily perform similarly on combined requests. Gemini~3.8 Flash and GLM-5.2 achieve Atomic Case pass rates of \GeminiAtomicCase{}\% and \GLMAtomicCase{}\%, respectively, but their Compositional Case pass rates are \GeminiCompositionCase{}\% and \GLMCompositionCase{}\%.

These results show that systems with similar aggregate performance on individual edits can differ in their ability to complete combined requests. We next examine performance across Word feature areas (\S\ref{sec:capability_profile}), then compare exactly linked Atomic and Compositional cases to assess whether success on individual operations is maintained when they are combined (\S\ref{sec:composition_burden}).
\begin{figure}[t]
\centering
\includegraphics[width=\linewidth]{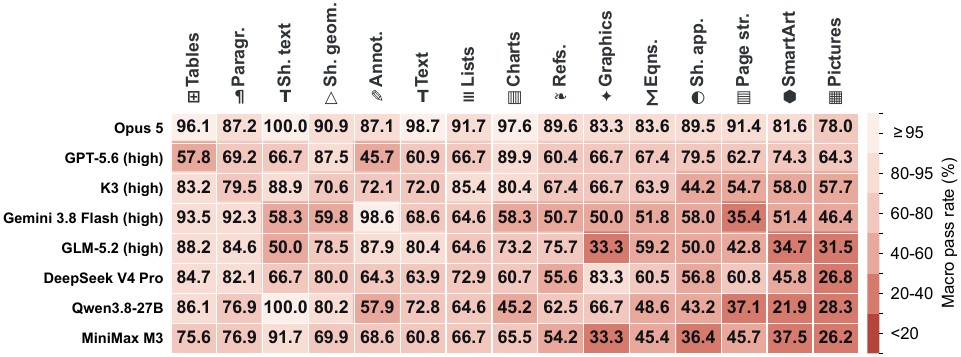}
\caption{Atomic macro-averaged pass rates (\%) across fifteen Word feature areas for eight agent systems. Each score averages command-level pass rates within the feature area, with missing predictions counted as failures. Columns are ordered by the mean across systems.}
\label{fig:capability-heatmap}
\end{figure}
\subsection{Atomic Capability Across Word Feature Areas}
\label{sec:capability_profile}
We analyze Atomic capability across the \nSubsystems{} Word feature areas introduced in \S\ref{sec:operation_space}. Figure~\ref{fig:capability-heatmap} shows macro-averaged pass rates under the \wordobj{} oracle for each feature area and system. \texttt{Tables} and \texttt{Paragraphs} achieve the highest mean pass rates, at $\FeatureTablesMean{}\%$ and $\FeatureParagraphsMean{}\%$, respectively, with strong performance on cell merging, row-height distribution, paragraph alignment, and indentation.

In contrast, \texttt{Pictures} has the lowest mean pass rate at $\FeaturePicturesMean{}\%$, followed by \texttt{SmartArt} ($\FeatureSmartArtMean{}\%$) and \texttt{Page Structure} ($\FeaturePageStructureMean{}\%$). These areas include image cropping, picture effects such as glow, bevel, and shadow, SmartArt node and formatting edits, and header/footer configuration. These results indicate that image and diagram editing and page-level configuration pose greater challenges than table and paragraph formatting in the evaluated Atomic tasks. We next examine whether success on individual operations carries over to compositional tasks that combine them.
\subsection{Atomic-to-Compositional Case Analysis}
\label{sec:composition_burden}
\begin{figure}[t]
\centering
\includegraphics[width=\linewidth]{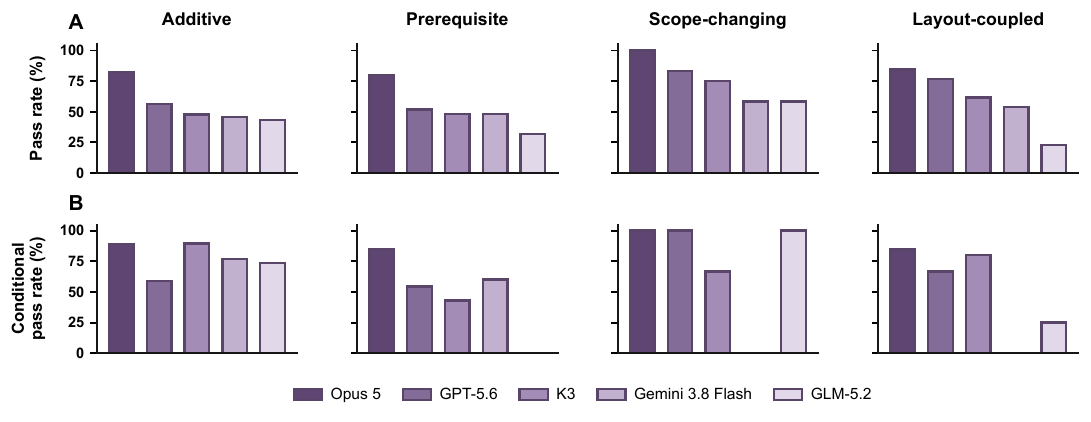}
\caption{Compositional case pass rates by class for five selected systems. (A) Case pass rates over all cases in each class. (B) Conditional case pass rates restricted to cases for which the same system passes all linked Atomic variants. Rates with fewer than five such cases are omitted. Appendix Table~\ref{tab:case-conditioning-denominators} reports the denominators for all eight systems.}
\label{fig:burden}
\end{figure}

We examine how success on individual Word operations relates to success when those operations are combined in one document.
All 96 Compositional cases carry the atomic-intent links defined in \S\ref{sec:atomic_intents}, covering 319 constituent intents per system.
Let $C$ denote a pass verdict on a Compositional case, and let $A$ denote the event that the same system passes all linked Atomic variants in isolation.
Figure~\ref{fig:burden} compares the case pass rate $P(C)$ with the conditional rate $P(C\mid A)$.
The condition selects a system-specific subset; Appendix Table~\ref{tab:case-conditioning-denominators} reports its size for each system and class.

\paragraph{Pass rates by composition class.}
Figure~\ref{fig:burden}A shows case pass rates across the four composition classes for five systems.
\textit{Scope-changing} tasks, in which earlier edits alter the targets of subsequent operations, have the highest pass rates for each system shown ($\SelectedScopeCaseMin{}$--$\SelectedScopeCaseMax{}\%$).
\textit{Layout-coupled} tasks, in which edits interact through document layout, show the largest variation across systems ($\SelectedLayoutCaseMin{}$--$\SelectedLayoutCaseMax{}\%$).
We examine these class-level differences further by conditioning on successful execution of the constituent Atomic tasks.

\paragraph{Conditioning on atomic-intent success.}
Figure~\ref{fig:burden}B reports $P(C\mid A)$, restricting the analysis to cases whose linked Atomic variants all pass in isolation.
For \textit{additive} tasks, conditional pass rates are higher than raw rates, but the difference varies across systems: $+\KimiAdditiveConditionalGain{}$ percentage points for K3 and $+\GPTAdditiveConditionalGain{}$ points for GPT-5.6.
The relationship is less uniform in other classes.
On \textit{layout-coupled} tasks, GPT-5.6 has a conditional pass rate of $\GPTLayoutConditionalCase{}\%$, compared with a raw rate of $\GPTLayoutCase{}\%$, while GLM-5.2 passes only two of its eight eligible cases ($\GLMLayoutConditionalCase{}\%$).
Thus, successful execution of all constituent operations in isolation does not ensure completion of the combined request.
As in the main results, missing outputs count as failures.

Appendix~\ref{app:intent_joint} complements this case-level view with intent-level outcomes.
For each intent whose Atomic case passes, we check whether the same intent also passes within the combined request, using intents whose embedded verdict is observed.
This transfer is lowest in \textit{layout-coupled} tasks.
These results suggest that operations that succeed in isolation are hardest to sustain when edits interact through document layout.

\subsection{Failure Mode Analysis}
\label{sec:failure_modes}
\begin{figure}[t]
\centering
\includegraphics[width=\linewidth]{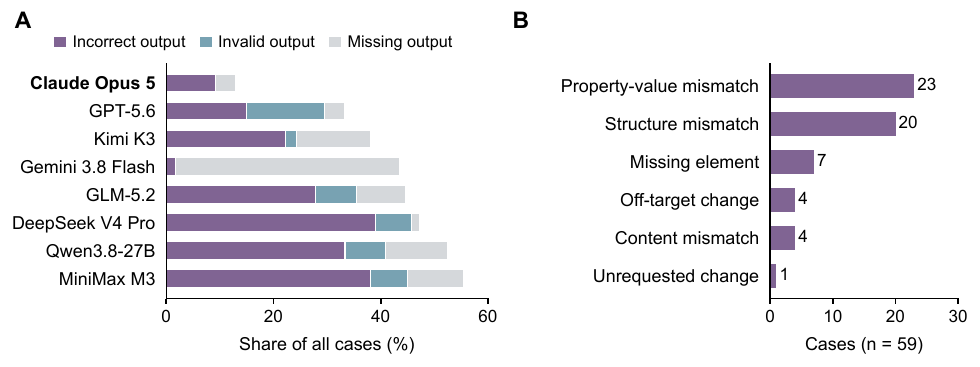}
\caption{Failure analysis under \wordobj{}.
\textbf{(A)} Output status of non-passing cases over \nCases{} cases per system.
\textbf{(B)} Mismatch types of the \nOpusIncorrectOutputs{} incorrect outputs of Opus~5 (bold in A), one type per case.}
\label{fig:failure-analysis}
\end{figure}
\paragraph{Output status.}
Each non-passing case is assigned to one of three outcomes (Figure~\ref{fig:failure-analysis}A).
Missing output means that no document was submitted, invalid output means that Word cannot open the document or it fails OOXML validation, and incorrect output means that a valid document fails a requested postcondition or preservation check.
For Opus~5, incorrect outputs account for \nOpusIncorrectOutputs{} of its \nOpusNonPassing{} non-passing cases, and no output is invalid.
For GPT-5.6, invalid and missing outputs together account for $\GPTInvalidMissingShare{}\%$ of all cases, while missing outputs alone account for $\GeminiMissingShare{}\%$ for Gemini~3.8 Flash.
These systems often fail before the requested edit can be evaluated.

\paragraph{Mismatch types of incorrect outputs.}
We inspected the evaluator diagnostics and saved OOXML of all \nOpusIncorrectOutputs{} incorrect outputs of Opus~5 and assigned each case to one of six types, namely property-value mismatch, structure mismatch, missing element, off-target change, content mismatch, or unrequested change (Figure~\ref{fig:failure-analysis}B).
Property-value mismatches are the largest type, accounting for $\OpusPropertyValueShare{}\%$ of the cases.
Of the rest, $\OpusStructureShare{}\%$ are structure mismatches, $\OpusMissingElementShare{}\%$ missing elements, $\OpusOffTargetShare{}\%$ off-target changes, $\OpusContentShare{}\%$ content mismatches, and $\OpusUnrequestedShare{}\%$ unrequested changes.
In property-value mismatches, the requested object is present with a different setting, such as a shadow preset or a footnote numbering format.
In picture-effect cases among structure mismatches, Word's object model reports the requested effect, but the saved OOXML lacks the image layer that Word writes for it.
Most missing elements are equation delimiters or matrices inserted with empty placeholders.
In the SmartArt text-fill task, Opus~5 fills the shape instead of its text, and in $\sin x$ and $\ln x$ it inserts an invisible function-application character (U+2061) that leaves the rendering unchanged.
These outcomes show that most incorrect outputs differ from the reference in settings or Word structure rather than omitting the requested element.

%% file: assets/tables/main_results.tex
\begin{table}[!t]
    \centering
    \caption{Pass rates (\%) on \benchname{} under the \wordobj{} oracle.
    \textbf{Case} weights cases equally, whereas \textbf{Macro} averages within each Atomic command or Compositional scenario and then across commands or scenarios (\S\ref{sec:setup}).
    Bold marks the best agent result in each column. Human reports the mean across three participants.}
    \label{tab:main-results}
    \small
    \setlength{\tabcolsep}{3pt}
    \begin{tabular}{@{}l@{\hspace{12pt}}
        *{2}{>{\raggedleft\arraybackslash}p{31pt}}@{\hspace{9pt}}
        *{2}{>{\raggedleft\arraybackslash}p{31pt}}@{\hspace{9pt}}
        *{2}{>{\raggedleft\arraybackslash}p{31pt}}@{}}
    \toprule
    \textbf{Agent system}
    & \multicolumn{2}{c}{\textbf{Atomic}}
    & \multicolumn{2}{c}{\textbf{Compositional}}
    & \multicolumn{2}{c}{\textbf{Overall}} \\
    \cmidrule(lr){2-3} \cmidrule(lr){4-5} \cmidrule(l){6-7}
    & Case & Macro & Case & Macro & Case & Macro \\
    \midrule
Claude Code (Opus 5) & \textbf{87.5} & \textbf{89.2} & \textbf{84.4} & \textbf{85.5} & \textbf{87.1} & \textbf{88.8} \\
Codex (GPT-5.6, high) & 67.6 & 68.2 & 61.5 & 60.3 & 66.7 & 67.3 \\
Kimi Code (K3, high) & 63.4 & 68.1 & 53.1 & 55.8 & 61.8 & 66.8 \\
OpenCode (Gemini 3.8 Flash, high) & 57.9 & 63.2 & 49.0 & 51.3 & 56.5 & 61.9 \\
ZCode (GLM-5.2, high) & 58.2 & 63.7 & 39.6 & 40.4 & 55.5 & 61.2 \\
OpenCode (DeepSeek V4 Pro) & 54.9 & 61.7 & 40.6 & 40.3 & 52.8 & 59.5 \\
Qwen Code (Qwen3.8-27B) & 50.4 & 56.4 & 31.2 & 32.4 & 47.5 & 53.9 \\
Mini Agent (MiniMax M3) & 47.3 & 54.9 & 29.2 & 30.6 & 44.5 & 52.4 \\
\midrule
Human (mean) & 85.2 & 86.7 & 69.4 & 68.4 & 82.8 & 84.8 \\
    \bottomrule
    \end{tabular}
\end{table}

%% file: sections/05_conclusion.tex
\section{Conclusion}

We introduced \benchname{}, a benchmark that evaluates document-editing agents by the runtime object properties and preserved document state that their edits produce in Word.
Its Atomic and Compositional tracks span the execution of individual operations and the coordination of multiple edits within a shared document state.
Across eight agent systems, many still struggle on Atomic tasks, performance varies across Word feature areas, and even the strongest system scores lower on Compositional tasks.
The atomic-intent and failure-mode analyses further expose operation-level limitations, composition-specific failures, and artifact-level issues.
Together, these results suggest that operation fidelity and compositional reliability are distinct requirements for document-editing agents.

\section{Limitations}
\label{sec:limits}
\paragraph{Synthetic controlled documents.}
\benchname{} is a controlled operation-fidelity benchmark, not a natural-document distribution or end-to-end office-workflow benchmark.
Its documents isolate Word-native edits under known preconditions, so results should not be read as measuring open-ended writing quality, user productivity, or arbitrary real-world document performance.

\paragraph{Coverage and portability.}
The current benchmark uses English, protocol-guided instructions; multilingual instructions and user-specific document conventions are outside the current evaluation scope.
The evaluator requires Windows and Microsoft Word because \benchname{} checks Word-resolved runtime state, including lists, fields, layout-dependent properties, and object effects, rather than only the saved \texttt{.docx} file structure.

\paragraph{Coverage-oriented, not frequency-weighted.}
The released set is a coverage-oriented instantiation of the in-scope verifiable operation space rather than a frequency-weighted sample of real user workloads.
Scores should therefore be read as capability-space coverage rather than expected accuracy on natural document-editing workloads.

\section{Broader Impacts}
\benchname{} enables more rigorous evaluation of document-editing agents by exposing failures that surface-text, screenshot, or coarse task-completion metrics miss, helping developers diagnose agents before deployment.
It is a capability benchmark rather than a deployment-safety guarantee, and high scores on controlled tasks do not imply safe behavior on out-of-distribution documents, long-horizon workflows, or adversarial inputs.

%% file: sections/99_appendix.tex
\section*{Supplementary Material}
\phantomsection
\pdfbookmark[0]{Supplementary Material}{supplement}
The supplement follows the benchmark from construction and verification to execution and additional analysis. Task examples and the complete command catalogue are collected at the end for reference.

\begingroup
\small
\renewcommand{\contentsname}{Appendix Contents}
\tableofcontents
\endgroup
\clearpage

\section{Benchmark Construction and Coverage}
\label{app:taxonomy}

\subsection{Command Selection and Scope}
\label{app:outofcoverage}

Table~\ref{tab:coverage-funnel} summarizes the command-selection stages used to construct the released Atomic set from Word's ribbon controls.
We start from Word ribbon controls~\citep{wordcontrols} and consolidate command-bearing entries into normalized semantic commands.
We then identify commands whose effects are state-changing and verifiable in Word-native state, yielding $\nVerifiableCommands{}$ verifiable commands across the three behavior families.
Within this verifiable pool, we define the current benchmark scope around document-editing operations; cross-application objects, form design, mail merge, macro/VBA authoring, multi-document workflows, automatic rewriting, and proofing are catalogued separately, yielding $\nInScopeCommands{}$ in-scope, benchmark-eligible commands.
We select the \nAtomicCommands{} released commands from this pool to cover all three behavior families and all \nSubsystems{} Word feature areas.
The remaining in-scope commands define an extension space for future releases.

\begin{table}[!htbp]
\centering
\caption{Command selection from Word ribbon controls to the released Atomic command set.}
\label{tab:coverage-funnel}
\small
\setlength{\tabcolsep}{6pt}
\begin{tabular}{@{}>{\raggedright\arraybackslash}p{0.36\linewidth}r>{\raggedright\arraybackslash}p{0.44\linewidth}@{}}
\toprule
\textbf{Stage} & \textbf{Count} & \textbf{Reduction} \\
\midrule
Ribbon control entries & \nRibbonControlEntries{} & \\
Normalized commands & \nNormalizedCommands{} & drop $202$ entries without a command identifier; deduplicate repeated placements \\
Verifiable state-changing commands & \nVerifiableCommands{} & exclude view, settings, clipboard, external, UI-structural, and selection commands ($1{,}419$; categories below) \\
In-scope commands & \nInScopeCommands{} & remove $158$ excluded-by-design commands (OLE, forms, mail merge, macro/VBA, master documents, AutoCorrect/AutoFormat, proofing) \\
\midrule
Released atomic commands & \nAtomicCommands{} & selected set spanning \nSubsystems{} feature areas; remaining in-scope commands define an extension space \\
\bottomrule
\end{tabular}
\end{table}

\paragraph{Outside the verifiable state-changing pool.}
These commands either affect state not stored in \texttt{.docx} or lack deterministic Word-native postconditions.
\begin{itemize}[leftmargin=*,itemsep=2pt]
\item \emph{View-state commands.} Zoom, ruler, navigation pane, gridlines, page color preview.
\item \emph{Document-settings commands.} Protect Document, Undo/Redo.
\item \emph{Clipboard operations.} Copy, Cut, Paste, Format Painter.
\item \emph{External-service commands.} Dictate, Translate, Copilot, Insert from Online, Smart Lookup.
\item \emph{UI-structural controls.} Menu openers, group labels, tab headers, dialog launchers without a state-changing action.
\end{itemize}

\paragraph{Outside the in-scope command set.}
\begin{itemize}[leftmargin=*,itemsep=2pt]
\item Embedded OLE / ActiveX objects (cross-application state).
\item Form controls (Word as form-design tool, not document editor).
\item Mail merge (multi-document workflow).
\item Macro / VBA authoring (program editing, not document editing).
\item Multi-document workflows (split, master/sub-documents).
\item AutoCorrect / AutoFormat (automatic rewriting outside an explicit command).
\item Proofing (spelling and grammar checking).
\end{itemize}

\subsection{Released-Command Selection}
\label{app:released_selection}
Starting from the $\nInScopeCommands{}$ in-scope commands, we include a command in the released Atomic set when it satisfies three construction requirements: the generator can construct an input document satisfying its precondition, Word automation can execute the command, and the resulting state is verifiable through Word-native postconditions.
Execution uses direct invocations for button commands, explicit parameter values for gallery commands, and documented object-model calls for dialog commands.
This yields the \nAtomicCommands{} released commands, spanning all \nSubsystems{} feature areas.
Table~\ref{tab:released-commands} lists the released commands by feature area, using Office Fluent UI command identifiers (\texttt{idMso}).

\subsection{Parameter Variants}
\label{app:construction_details}
A single case per command would conflate command-level ability with familiarity with one particular parameter value.
To ensure that scores reflect general command execution rather than memorization of specific parameter settings, the generator creates variants over the parameter space of each command.
For parameterized commands such as galleries, dropdowns, and dialogs, the generator enumerates the available options through Word's object model and creates cases that span up to eight options per command.
Toggle commands are tested in both their on and off states, and other commands typically have one case.
The same Reference Executor produces $d^*$ and exports the corresponding oracle specification for each variant, so adding variants requires no manual document authoring or labeling.

\subsection{Human Review of Benchmark Cases}
\label{app:quality_control}
Three annotators experienced with Microsoft Word independently reviewed every released benchmark case for instruction clarity, input-document validity, and reference-document correctness.
Cases flagged during review were manually adjudicated, and affected instructions and reference documents were corrected and revalidated before release.
We thank TELUS Digital for supporting the human annotation studies.

\FloatBarrier

\section{Word-Native Verification and Validation}
\label{app:verification}
The oracle checks the requested edit and the preservation conditions specified for each case. This section details the object--property coverage and verification schemas, then compares evaluator precision, false-positive rates, and false-negative rates.

\subsection{Object--Property Coverage}
\label{app:object_coverage}

Word exposes document objects and their properties through its object model~\citep{wordobjectmodel}; Office Open XML specifies the saved document format~\citep{ooxml}. The following table summarizes the Word objects and properties covered by the oracle, organized by Word feature area. Each case's oracle specification defines the target postconditions and preservation checks applied to that case.
\input{assets/tables/object_property_coverage}

\subsection{Target Postconditions and Preservation}
\label{app:verification_schemas}

\begin{table}[!htbp]
\centering
\caption{Representative verification schemas, target postconditions, and preservation categories from the benchmark design. Each case applies the checks defined by its oracle specification; not every listed invariant applies to every case.}
\label{tab:preservation}
\small
\setlength{\tabcolsep}{5pt}
\renewcommand{\arraystretch}{1.2}
\begin{tabular}{@{}>{\raggedright\arraybackslash}p{0.24\linewidth} >{\raggedright\arraybackslash}p{0.36\linewidth} >{\raggedright\arraybackslash}p{0.30\linewidth}@{}}
\toprule
\textbf{Verification schema} & \textbf{Target postcondition} & \textbf{Preservation invariants} \\
\midrule
Run / character formatting & \texttt{Range.Font.*} (weight, style, size, color, highlight, text effects) on the target range & Non-target (``control'') text content and formatting; paragraph count \\
\addlinespace
Paragraph formatting & \texttt{ParagraphFormat.*} (alignment, indents, line spacing, space before/after, borders) & Paragraph text; non-target paragraph properties; paragraph count \\
\addlinespace
Lists & \texttt{ListFormat} (list string, list level) on the target paragraphs & Paragraph text; non-target list state; paragraph count \\
\addlinespace
Shape / picture geometry \& effects & \texttt{Shape.\{Left, Top, Width, Height, Rotation, ZOrderPosition\}}, wrap type, fill / line / shadow / glow / reflection & Document structure counts (shape, inline-shape, footnote, section); non-target shapes \\
\addlinespace
Tables & Row and column counts, cell content, borders, style & Table dimensions; non-target cells \\
\addlinespace
Fields \& references (footnote, endnote, TOC, citation) & Field count, resolved content, reference position & Body text when the field is the sole mutation \\
\addlinespace
Headers / footers & Header/footer story content, field count & Document body text; section count \\
\bottomrule
\end{tabular}
\end{table}

\FloatBarrier

\subsection{Evaluator Validation}
\label{app:oracle_validity}

\paragraph{Positive control.}
We submit the reference document of each of the 303 cases in place of the agent output. Word Object accepts all 303.

\paragraph{Human reference.}
We compare evaluator judgments with the majority of three human evaluators on 303 agent-generated Word documents.
Each human evaluator judged the agent output against the instruction by opening it alongside the input document in Word, without seeing any evaluator verdict, and could open the reference document on request.
We exclude 41 documents that Word cannot open or that fail XML or package-structure checks, and four documents with no binary majority.
The comparison set contains 258 documents, 167 labeled Pass and 91 labeled Fail.
The box below reproduces the instructions given to the human evaluators.
\begin{tcolorbox}[breakable, colback=black!4!white, colframe=black!55!white, fonttitle=\bfseries\small, title={Human evaluator instructions}]
\footnotesize
\begin{verbatim}
# WordEval - Human Judging Package

You will judge whether an AI model correctly edited a Word document.
Your labels help validate WordEval's automated checking against human
judgment.

For each case you see an instruction and download two documents - the
Original (before the edit) and a Candidate (the model's attempt). You
open both in Word and decide, independently: did the Candidate
correctly carry out the instruction? You are never shown any automated
verdict - your own judgment is the point.

Some instructions bundle several sub-edits. Those show a checklist: you
judge each part separately, then give an overall verdict.

[... setup steps omitted ...]

## What You Need
- Windows with Microsoft Word - you judge the real .docx in Word, not a
  picture. A flat image hides differences that matter (a real footnote
  vs a typed superscript, a real field vs typed text, styles, page
  breaks vs blank lines).

## How to Judge a Case
1. Read the instruction shown on screen.
2. Download both documents and open them in Microsoft Word.
3. Compare before -> after. Helpful Word tools: Home -> Show/Hide
   (show marks), Reveal Formatting (Shift+F1).
4. Answer:
   - Does the Candidate correctly perform the instruction? - Pass /
     Fail / Unsure
   - How confident are you? - High / Medium / Low
   - Notes (optional, but please add them when unsure or when
     something is subtle).
5. Click Submit & next. [...]

How to decide:
- Judge against the instruction, not against one specific output - a
  different but valid way still passes. There is no single required
  result.
- Missing the required change, wrong target, wrong element type, or
  destroying other content -> fail.
- Pick unsure only when you genuinely cannot tell even after opening
  both documents in Word.

### Multi-part cases
When the instruction bundles several sub-edits, judge each checklist
row on its own merits: did that sub-edit happen correctly? Then give
the overall verdict separately - it is not automatically "all parts
pass": parts can interfere with each other, and unrelated content may
have been damaged.

### Stuck?
If you genuinely cannot tell what a correct result should look like
(e.g. an obscure Word feature), expand "Stuck? Reveal one correct
example" to download a Reference.docx - one example of a correct
output. Use it sparingly, and keep judging by the instruction: a
result that differs from the example can still pass. You will not
need it for most cases.

[... file-return steps omitted ...]

## Time
There is no time limit - take your time on each case. (A timer runs
silently for analysis; it is never shown and has no cap.) If a case
feels broken or impossible after a reasonable effort, mark it unsure
with a note and move on.

[... troubleshooting omitted ...]
\end{verbatim}
\end{tcolorbox}

\paragraph{Compared evaluators.}
Besides Word Object, we evaluate Word Revision~\citep{wordcompare}, SSIM~\citep{ssim} between RGB renderings of the reference and submitted documents at thresholds of 0.99 and 0.999, and an agent judge.
The agent judge runs GPT-6 Astra through Codex CLI with high reasoning effort and document-analysis tools. It sees the instruction, the input document, and the submitted document, but not the reference document, the human judgments, or Word Object results.

\paragraph{Metrics.}
Pass is the positive class. Precision is the fraction of accepted outputs that the human majority labels Pass.
The false-positive rate (FPR) is the fraction of majority-Fail outputs accepted, and the false-negative rate (FNR) is the fraction of majority-Pass outputs rejected.

\input{assets/tables/human_oracle_audit}

\paragraph{Results.}
Word Object accepts 5 of the 91 majority-Fail outputs and rejects 28 of the 167 majority-Pass outputs, for 96.5\% precision, 5.5\% FPR, and 16.8\% FNR (Table~\ref{tab:human-oracle-audit}).
Word Revision reaches a similar precision of 97.1\% but rejects 100 majority-Pass outputs, an FNR of 59.9\%.
In the comparison set, it reports every SmartArt output as one shape deletion and one shape insertion, whether or not the edit is correct. It therefore rejects all 13 majority-Pass SmartArt outputs.
At SSIM $\geq 0.99$, 41 of the 91 majority-Fail outputs pass, an FPR of 45.1\%.
Raising the threshold to 0.999 lowers the FPR to 7.7\% but raises the FNR from 16.2\% to 27.5\%. Word Object is lower on both rates.
In one header page-numbering task, the submitted document passes even the 0.999 threshold, yet both the human majority and Word Object reject it. Word Object reports a mismatch in the header style.
The agent judge rejects fewer majority-Pass outputs than Word Object (FNR 12.0\% versus 16.8\%) but accepts 11 majority-Fail outputs, more than twice the FPR of Word Object (12.1\% versus 5.5\%).

\clearpage
\section{Evaluation Setup and Prompts}
\label{app:agent_setup}
We evaluate the eight agent systems listed in Table~\ref{tab:main-results}.
\subsection{Agent Inputs and Task Prompt}
\label{app:agent_prompt}
Agents receive the task instruction and input document and are asked to return \texttt{pred.docx} with trajectory documentation. Reference documents and oracle files are available only to the evaluator. Figure~\ref{fig:task-prompt} illustrates the task-file protocol. CLI wrappers and model-specific effort settings vary across systems.
\begin{figure}[!htbp]
\begin{tcolorbox}[colback=black!4!white, colframe=black!55!white, fonttitle=\bfseries\small, title={Per-case prompt}]
\footnotesize\ttfamily
Please follow @TASK\_DESCRIPTION\_per\_task.md to complete the task. Only access files in the current directory.
\end{tcolorbox}
\begin{tcolorbox}[colback=black!4!white, colframe=black!55!white, fonttitle=\bfseries\small, title={Task description (\texttt{TASK\_DESCRIPTION\_per\_task.md}), abridged}]
\footnotesize
\begin{verbatim}
# Word Command Benchmark - Task Description

## Overview
This benchmark evaluates your ability to execute Microsoft Word
formatting and editing commands on documents. Each test case provides
an input Word document and a natural language instruction describing
the modification to perform.

## Task Structure
The current directory contains a single test case:
  doc_in.docx       # Input document
  instruction.txt   # Task instruction
  pred.docx         # YOUR OUTPUT (to be created)
  method.md         # YOUR METHOD (full reproducible trajectory)

## Your Task
1. Read the instruction.txt file to understand what modification
   needs to be made
2. Open the doc_in.docx file (input document)
3. Execute the instruction on the document
4. Save the result as pred.docx in the current directory
5. Document your complete trajectory in method.md

## Method Documentation Requirements
# Status
STATUS: [SUCCESS|PARTIAL|FAILED]
REASON: [Optional: explain if partial or failed]
# Approach / # Step-by-Step Trajectory / # Complete Implementation

### Status Definitions
- SUCCESS: Task completed as instructed
- PARTIAL: Task attempted but may not be fully correct
- FAILED:  Could not complete the task

You can use any method you prefer.
[... four worked examples (python-docx, VBA, Word GUI, other tools
 such as Office.js, OpenXML SDK, Apache POI, pandoc) omitted ...]
\end{verbatim}
\end{tcolorbox}
\caption{Task-file protocol with an abridged task description.}
\label{fig:task-prompt}
\end{figure}

\FloatBarrier
\subsection{Agent Systems and Run Settings}
\label{app:run_settings}
Qwen Code uses a 50-turn budget and OpenCode with Gemini a 50-step budget, each with a 600-second per-case limit. Failed runs and missing outputs remain in the 642-case denominator. All outputs were evaluated with Microsoft Word \EvalWordBuild{} on Windows~11 virtual machines (OS build \EvalOSBuild{}) with Python \EvalPythonVersion{}. Opus~5 uses a Copilot compatibility route for the Stack Object case. Table~\ref{tab:agent-versions} reports packages, available versions, and model identifiers.
\input{assets/tables/agent_versions}
\FloatBarrier
\subsection{Human Execution Protocol}
\label{app:human_execution}
Three participants (A01--A03) each perform the same 642 editing tasks in Microsoft Word, through a local browser page that presents one task at a time with its instruction and input document. Tasks are grouped into bundles, and the Atomic tasks in a bundle precede the Compositional task that combines them. A practice task precedes the benchmark tasks and is excluded from analysis. For each task, the participant downloads the input document, carries out the instruction in Word, and uploads the saved document. The instructions given to participants, reproduced in the box below, specify which tools may be used. The final output submitted for each task is scored with the same evaluator as agent outputs. These execution results are distinct from the human judgments of agent outputs in Appendix~\ref{app:oracle_validity}. Table~\ref{tab:human-execution} reports individual Case and Macro pass rates, and the Human row in the main table averages the three participant-specific rates.
\input{assets/tables/human_execution}
\begin{tcolorbox}[breakable, colback=black!4!white, colframe=black!55!white, fonttitle=\bfseries\small, title={Human execution instructions, abridged}]
\footnotesize
\begin{verbatim}
# WordEval Human Annotation

This task evaluates whether a human annotator can execute a Microsoft
Word editing instruction on a document, and validates the WordEval
benchmark cases (instruction clarity + Word Object oracle correctness).
Each task contains one input Word document and one natural-language
instruction.

Cases are organized into task bundles. Within each bundle, complete the
atomic tasks first, then complete the corresponding compositional task.

Each task has two stages:
1. Stage 1: Word-only baseline - complete the task using only memory,
   ribbon browsing, and in-document text search (no oracle).
2. Stage 2: Assisted attempt - optional, after Stage 1. Use external
   help to attempt the task again, starting from a fresh copy of the
   input document.

[... setup, onboarding, and step-by-step interface instructions omitted ...]

Stage 1 - Allowed
- Memory and ribbon/menu browsing
- Right-click menus and keyboard shortcuts
- Ctrl+F for in-document text search

Stage 1 - Forbidden
- Search box at the top of the Word window (the bar above the ribbon
  labeled "Search" or "Tell me what you want to do", with a magnifying
  glass icon). Typing into it suggests commands or invokes Copilot.
- Copilot panel (the AI sidebar accessible from the ribbon)
- F1 Help when it opens docs.microsoft.com or any web page
- Web search, ChatGPT, Claude, or any other AI assistant
- Scripts, macros, VBA, WordAgent, or programmatic editing

Stage 1 measures what a Word user can do from memory and ribbon
browsing alone, matching the agent's no-oracle condition.

Stage 2 - Allowed
- Everything Stage 1 allows, plus:
- Search box at the top of the Word window ("Search" / "Tell me")
- Word's Copilot panel (for explanations only)
- F1 Help and docs.microsoft.com
- Web search, YouTube tutorials, Stack Overflow
- ChatGPT / Claude / external Copilot as a tutor (ask, read, then
  perform the clicks yourself)

Stage 2 - Forbidden
- AI directly editing the docx (no AI-generated macros, VBA, scripts,
  or XML run on the document)
- Pasting AI-generated final content into the docx as a shortcut
- Any agent or script that performs the edit on your behalf

AI may explain, not execute. You must perform the final clicks in Word
yourself.

Compare to Gold (Submit step)
Before final submission, open gold.docx (the reference output) in Word
alongside your final uploaded result (pred_stage2.docx if you
continued, otherwise pred_stage1.docx), and rate human_matches_gold.
[...]

Instruction Specification Severity (0-3)
Rate from the perspective of another reasonable Word user, not your
own struggle. Anchors:
- 0 - clear, all reasonable annotators converge on the same edit
- 1 - minor interpretation needed, a reasonable default exists
- 2 - important detail missing or multiple valid interpretations
- 3 - too ambiguous to execute, or contradicts the docx

human_matches_gold (Binary)
- match - the targeted edits from the instruction are realized in your
  output as in gold
- not_match - at least one targeted edit is missing or applied
  differently

[... status and confidence definitions omitted ...]

Time Policy
There is no time limit. Take your time on each case. If a case feels
broken or impossible after a reasonable effort (~10 min), mark it as
failed and move on.
\end{verbatim}
\end{tcolorbox}
\clearpage
\subsection{Code and Data Availability}
\label{app:artifact}
The released \benchname{} artifact is organized around a canonical benchmark directory with separate Atomic and Compositional tracks.
Each track contains agent-visible task records, input \texttt{.docx} files, evaluator-only reference documents, oracle metadata, and task-level taxonomy metadata.
The agent-visible records specify only the task identifier, natural-language instruction, and input document path; the evaluator-only records specify the corresponding reference document and oracle specification used to score predictions.
This separation mirrors the evaluation protocol in Section~\ref{sec:task_formulation}: agents receive the input document and instruction, while reference documents and oracle metadata are used only by the evaluator.

The artifact also includes the benchmark generator, Word Object evaluator, scripts for producing agent-visible task folders and scoring prediction files, a README, a data card, license files, and Croissant metadata.
The primary evaluator requires Windows, Microsoft Word, and Python COM automation, matching the environment assumptions in Section~\ref{sec:setup}.
Code is released under the MIT License, and the benchmark data are released under CDLA-Permissive-2.0.
Code and data are available at
\url{https://aka.ms/word-eval}.
The data card and Croissant metadata document the dataset schema, intended use, limitations, privacy properties, licensing, and responsible-release metadata.

\FloatBarrier

\section{Additional Compositional Analyses}
\label{app:intent_burden}

The main-text analysis (\S\ref{sec:composition_burden}) treats each compositional case as a single binary outcome, so the whole case either passes or fails.
Every released compositional case (96 cases / 319 atomic intents across the four classes) declares the atomic intents it contains via \texttt{spec.composition.atomic\_intents}, and per-intent verdicts are extracted from the recorded verifier diagnostics when available.
This appendix uses those per-intent verdicts to decompose outcome patterns that case-level outcomes collapse.
These are operation-limited observations where the embedded intent fails along with its referenced Atomic case, transfer failures where a referenced Atomic pass verdict does not carry over to the corresponding embedded atomic intent, and end-state coincidences where a compositional final state satisfies an intent despite the corresponding Atomic failed verdict.
Operation, feature area, and depth breakdowns are reported as secondary diagnostics after this intent-level view.

\subsection{Intent-Level Outcome Decomposition}
\label{app:intent_protocol}
\label{app:intent_joint}

\paragraph{Definitions.}
For each declared intent $i$, $A_i$ is the verdict of its referenced Atomic case in isolation, and $C_i$ is the verdict of the same intent inside the compositional output.
The transfer rate $\Pr(C_i \mid A_i)$ is the share of intents passing in isolation that also pass inside the composition.
Each pair of $A_i$ and $C_i$ verdicts falls into one of four cells.
AC marks an intent that passes in both settings.
A$\bar{C}$ marks a \emph{transfer failure}, where the Atomic case passes but the embedded intent fails.
$\bar{A}\bar{C}$ marks an \emph{operation-limited} observation, where both fail.
$\bar{A}$C marks an \emph{end-state coincidence}, where the compositional output satisfies the intent's final-state check although the Atomic case fails.
The cells describe outcomes, not their causes.

\paragraph{Coverage.}
The eight systems yield $8 \times 319 = \nIntentDeclared{}$ declared intent observations.
Embedded verdicts come from the verifier's per-intent diagnostics whenever a prediction document exists and its diagnostics cover the intent.
Two cases that record their checks in other diagnostic structures are mapped explicitly.
Each intent verdict thus reflects only its own check.
Together with the Atomic verdicts, these yield \nIntentPairs{} observed pairs ($66.7\%$), from $36.1\%$ to $92.5\%$ across systems, with the remainder lacking an embedded or a binary Atomic verdict.
All intent-level rates are computed over these observed pairs.

\input{assets/tables/composition_outcome_decomposition}

\paragraph{Composition classes.}
Table~\ref{tab:intent-outcomes} pools the four cells across the eight systems by composition class.
Its transfer-failure column is A$\bar{C}$ over all Atomic-pass pairs, its operation-limited column is the share of $\bar{A}\bar{C}$ among all pairs, and its coincidence column is the share of $\bar{A}$C among all pairs.
Layout-coupled cases have the highest transfer-failure rate, $28.1\%$. Additive and prerequisite cases have the largest operation-limited shares, $9.7\%$ and $8.2\%$.
In layout-coupled cases, 45 of the 48 failed embedded intents passed in isolation.
In additive and prerequisite cases, 113 of 195 and 18 of 28 failed embedded intents also failed in isolation.
Embedded failures in layout-coupled cases are therefore mostly transfer failures, whereas more than half of those in additive and prerequisite cases are operation-limited.

\subsection{Intent Transfer by System}
\label{app:intent_class}

\begin{figure}[!htbp]
\centering
\includegraphics[width=\linewidth]{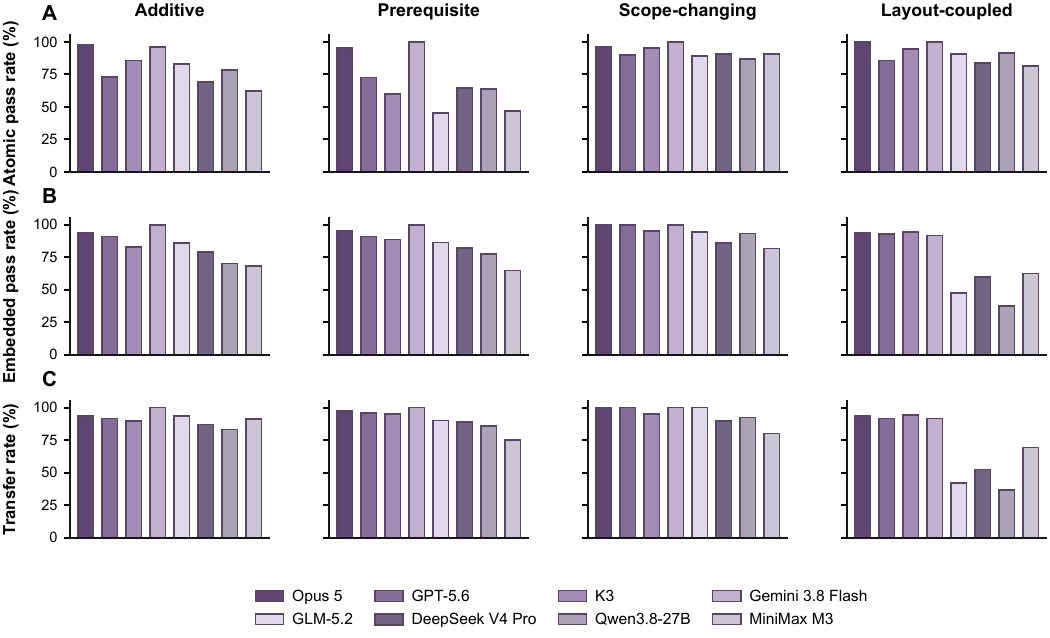}
\caption{Intent-level rates by system and composition class.
\textbf{(A)} Atomic pass rate $\Pr(A_i)$.
\textbf{(B)} Embedded pass rate $\Pr(C_i)$, measured inside the compositional output.
\textbf{(C)} Transfer rate $\Pr(C_i \mid A_i)$, among intents whose Atomic case passes in isolation.
Rates use observed pairs, and Table~\ref{tab:case-conditioning-denominators} lists the values.}
\label{fig:intent-transfer-all}
\end{figure}

\paragraph{Agent systems.}
Figure~\ref{fig:intent-transfer-all} reports $\Pr(A_i)$, $\Pr(C_i)$, and the transfer rate $\Pr(C_i \mid A_i)$ across the four classes for all eight agent systems.
This is the intent-level counterpart to the case-level analysis in Figure~\ref{fig:burden}.
Panel A gives the Atomic pass rate, panel B asks whether embedded intents pass in compositional outputs, and panel C asks the same question only for intents whose referenced Atomic case passes in isolation.
Class differences vary across systems.
Opus~5, GPT-5.6, Gemini~3.8 Flash, and K3 transfer at least $89.9\%$ of their Atomic-pass intents in every class, while the other four systems have their lowest transfer on layout-coupled cases.
Because intents without an observed embedded verdict are excluded, some system--class cells have small denominators (Table~\ref{tab:case-conditioning-denominators}).

For the case-level conditioning used in Figure~\ref{fig:burden}, Table~\ref{tab:case-conditioning-denominators} reports the case-level rates and the all-components-pass denominators for all eight systems, together with their intent-level rates.
Its intent-level columns place $\Pr(A_i)$ and $\Pr(C_i)$ side by side, and their difference is the pass gap.
On layout-coupled cases, the pass gap is largest for the same four systems, suggesting composite-intent loss that is not fully explained by atomic-component verdicts on the same operations.
Section~\ref{app:trivial_pass} examines cells with negative gaps.

\input{assets/tables/composition_case_conditioning}

\paragraph{Closed- and open-weight systems.}
Among the evaluated agent systems, those using open-weight models (K3, MiniMax~M3, Qwen3.8-27B, GLM-5.2, DeepSeek V4 Pro) have lower values than those using closed-weight models (Opus~5, GPT-5.6, Gemini~3.8 Flash) on both the embedded pass rate and the transfer rate (Figure~\ref{fig:intent-transfer-all}, Table~\ref{tab:burden-tiers}).
A paired case-cluster bootstrap~\citep{efronbootstrap} with 2,000 resamples of shared cases estimates positive closed-weight versus open-weight transfer differences in all four classes, $+5.3$\,pp for additive (95\% CI $[1.2, 10.2]$), $+8.2$\,pp for prerequisite ($[0.0, 18.1]$), $+7.7$\,pp for scope-changing ($[0.0, 18.4]$), and $+35.7$\,pp for layout-coupled cases ($[19.5, 53.7]$).

\input{assets/tables/composition_transfer_significance}

\FloatBarrier
\subsection{Breakdown by Operation and Word Feature Area}
\label{app:intent_op_sub}

\input{assets/tables/composition_transfer_by_operation}

Table~\ref{tab:burden-operation} ranks atomic operations by intent-level transfer rate, pooled across the eight agent systems and restricted to operations with at least 15 observed pairs.
Apart from shadow and reflection effects ($29.4$--$44.4\%$), the lowest-transfer operations are geometric/spatial primitives (\texttt{rotate\_angle} $47.2\%$, \texttt{flip\_direction} $58.8\%$, \texttt{align\_command} $72.6\%$), followed by list and icon-insertion primitives.
This is consistent with the pattern by composition class.
Layout-coupled and certain additive bundles share some of the same problematic primitives, suggesting that the underlying \emph{spatial coordination} that those classes over-represent contributes to low transfer.

\input{assets/tables/composition_transfer_by_subsystem}

Table~\ref{tab:burden-subsystem} groups transfer by Word feature area, the axis of Figure~\ref{fig:capability-heatmap}.
\texttt{Pictures} ($+12.5$\,pp), \texttt{Lists} ($+10.4$\,pp), and \texttt{Shape geometry} ($+10.1$\,pp) have the largest pass gaps.
\texttt{Text} and \texttt{Equations} have negative gaps ($-19.3$\,pp and $-30.7$\,pp) because end-state coincidences outnumber transfer failures in these areas.
Section~\ref{app:trivial_pass} examines these coincidences.

\subsection{End-State Coincidence Audit}
\label{app:trivial_pass}

\input{assets/tables/composition_trivial_pass_examples}

Intents can also pass more often inside compositions than in isolation.
The pass gap $\Pr(A_i) - \Pr(C_i)$ is negative in prerequisite cases for six of the eight systems, and GPT-5.6 has negative gaps in every class, from $-7.1$ to $-18.2$\,pp.
In these cells, end-state coincidences outnumber transfer failures.

The $\bar{A}$C cell -- Atomic-failed but composite-intent-passed -- accounts for 193 of 1,701 observed pairs ($11.3\%$), of which 121 ($62.7\%$ of the end-state coincidence set, $7.1\%$ of all observed pairs) appear in cases that also pass at case level.
In a composite, another unit or a no-op can sometimes satisfy the same end-state predicate (e.g., \texttt{ungroup\_command} passes when no group exists).
The transfer rate $\Pr(C_i \mid A_i)$ is computed over Atomic-pass pairs, so the $\bar{A}$C cell does not enter it, and the case-level metric $P(C)$ counts these cases as passes.

The most frequent operations among these coincidences are \texttt{insert\_equation} (17), \texttt{drop\_cap} (12), \texttt{outline\_weight} (10), \texttt{outline\_color} (8), and \texttt{rotate\_angle} and \texttt{table\_style} (7 each).
Table~\ref{tab:burden-trivial} shows one example from a passing case for each of these operations and for \texttt{ungroup\_command}.

\clearpage
\subsection{Composition Depth}
\label{app:burden_depth}

\begin{figure}[!htbp]
\centering
\includegraphics[width=\linewidth]{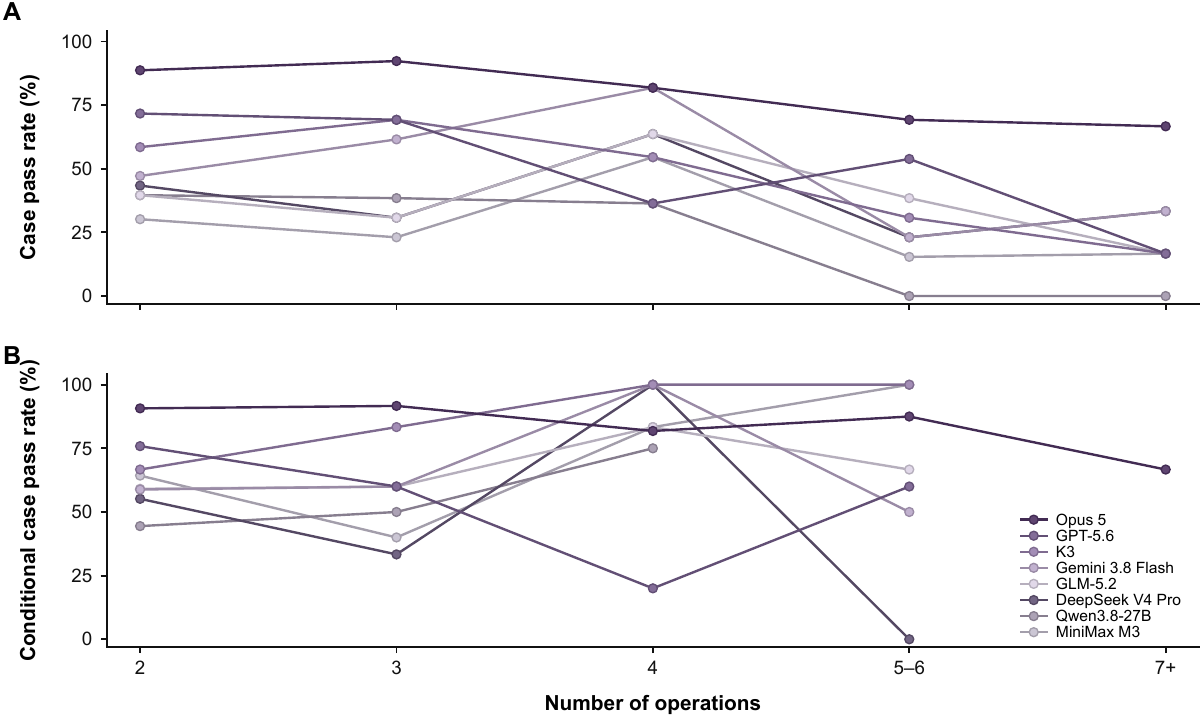}
\caption{Case-level pass rates by composition depth for the eight systems.
\textbf{(A)} $P(C)$ over all assigned cases, counting missing outputs as failures.
\textbf{(B)} $P(C\mid A)$ over cases whose linked Atomic variants all pass, omitted where no case is eligible.
Depth buckets hold 53, 13, 11, 13, and 6 cases at $k=2$, $3$, $4$, $5$--$6$, and $k\geq7$, and the $k\geq7$ bucket has a mean depth of $10.2$.}
\label{fig:burden-by-k}
\end{figure}

\begin{figure}[!htbp]
\centering
\includegraphics[width=\linewidth]{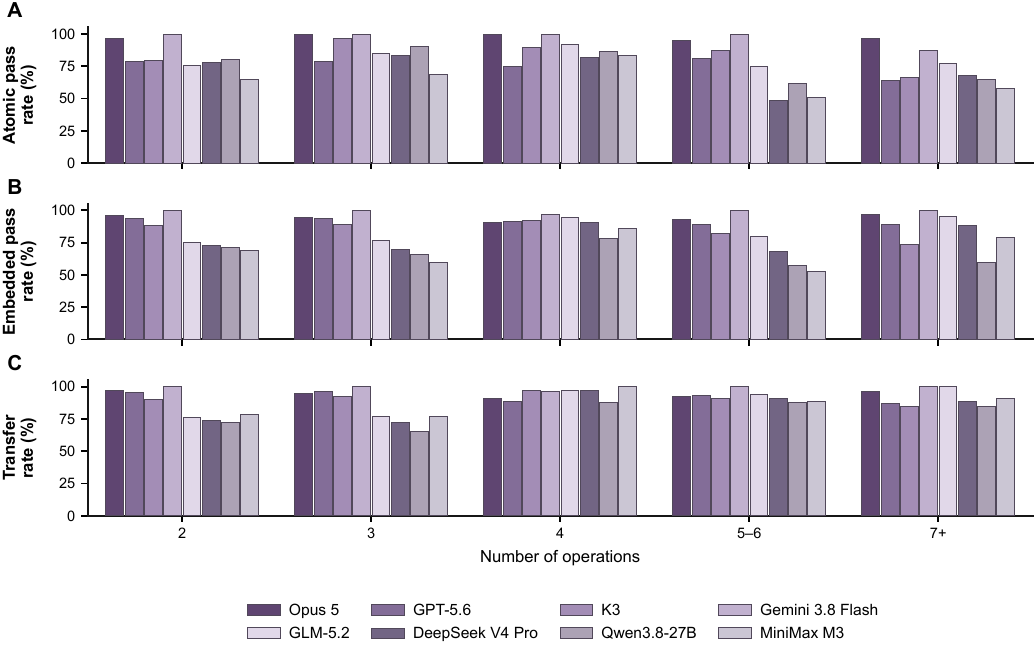}
\caption{Intent-level rates by composition depth for the eight systems.
\textbf{(A)} Atomic pass rate $\Pr(A_i)$.
\textbf{(B)} Embedded pass rate $\Pr(C_i)$.
\textbf{(C)} Transfer rate $\Pr(C_i \mid A_i)$.
Rates use observed pairs.}
\label{fig:intent-transfer-by-k}
\end{figure}

Composition depth $k$ is the number of atomic intents in a compositional case.
The case-level pass rate of Opus~5 is $88.7\%$ at $k{=}2$ and $92.3\%$ at $k{=}3$, then falls to $81.8\%$ at $k{=}4$, $69.2\%$ at $k{=}5$--$6$, and $66.7\%$ at $k{\geq}7$ (Figure~\ref{fig:burden-by-k}).
Its per-intent transfer stays between $90.9\%$ and $96.8\%$ in every bucket, including $k{\geq}7$ (Figure~\ref{fig:intent-transfer-by-k}).

\FloatBarrier
\section{Task Examples and Complete Command Catalogue}
\label{app:catalogue}
This section collects the task instructions, input-document examples, and command enumeration supporting Appendix~\ref{app:taxonomy}.

\subsection{Representative Case Examples}
\label{app:case-examples}
\begin{figure}[!htbp]
\centering
\begin{tcolorbox}[colback=black!4!white, colframe=black!55!white, fonttitle=\bfseries\small, title={Instruction}]
\footnotesize\itshape
Distribute all three rectangles evenly horizontally with equal spacing between them.
\end{tcolorbox}
\vspace{0.5em}
{\small\bfseries Input document (\texttt{doc\_in.docx})}\par
\vspace{0.3em}
\includegraphics[width=.9\linewidth,height=.33\textheight,keepaspectratio]{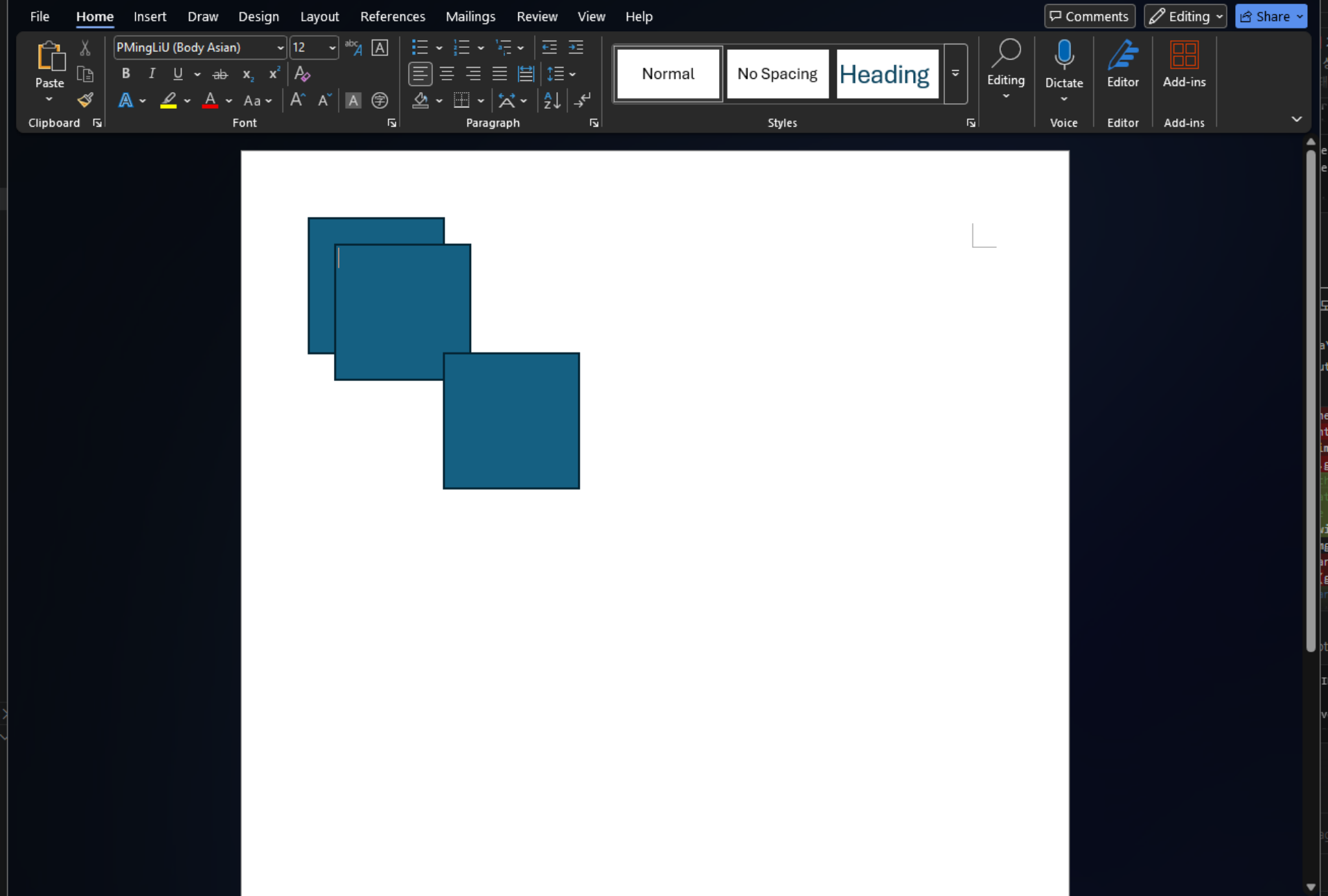}
\caption{Instruction and input document shown to agents for a representative atomic case (\texttt{AlignDistributeHorizontally}, Shape geometry feature area). The input page render is cropped to the content region and presented in a Windows Word application frame.}
\label{fig:atomic-case-input-instruction}
\end{figure}

\begin{figure}[!htbp]
\centering
\begin{tcolorbox}[colback=black!4!white, colframe=black!55!white, fonttitle=\bfseries\small, title={Instruction}]
\footnotesize\itshape
Apply square wrapping to the cross shape (text wraps around the cross's rectangular bounding box), then position it at the top-center of the page, using the Position preset (within default page margins).
\end{tcolorbox}
\vspace{0.5em}
{\small\bfseries Input document (\texttt{doc\_in.docx})}\par
\vspace{0.3em}
\includegraphics[width=.9\linewidth,height=.33\textheight,keepaspectratio]{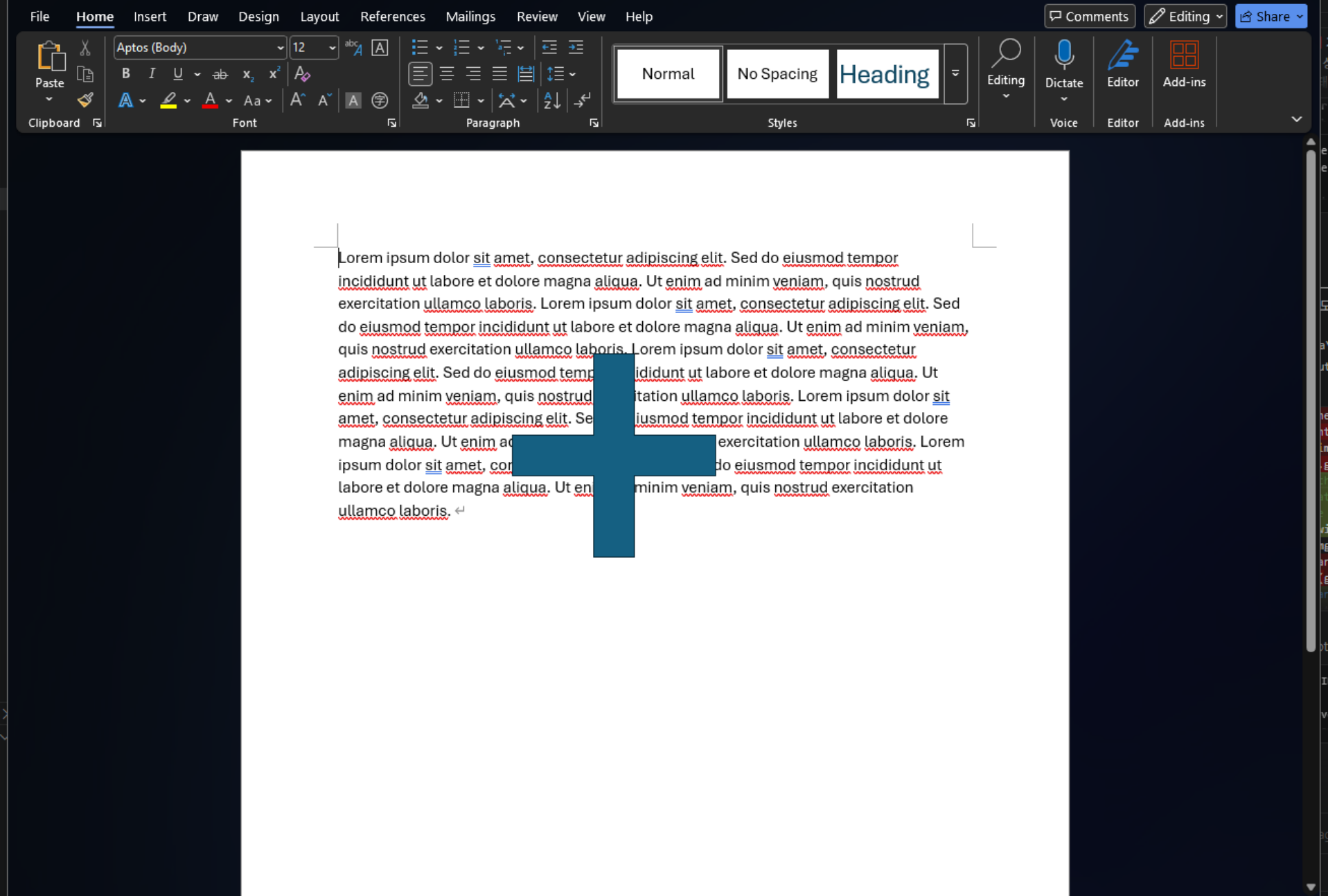}
\caption{Instruction and input document shown to agents for a representative layout-coupled compositional case (\texttt{TextWrapGallery} followed by \texttt{PicturePositionGallery}). The input page render is cropped to the content region and presented in a Windows Word application frame.}
\label{fig:compositional-case-input-instruction}
\end{figure}

Tables~\ref{tab:atomic-case-examples} and~\ref{tab:compositional-case-examples} give representative instructions shown to agents for single-operation Atomic cases and multi-operation Compositional cases, respectively. Figures~\ref{fig:atomic-case-input-instruction} and~\ref{fig:compositional-case-input-instruction} show, for one released case per track, the exact instruction together with a rendered preview of the input document.

{
\small
\setlength{\tabcolsep}{4pt}
\renewcommand{\arraystretch}{1.13}
\begin{longtable}{@{}>{\raggedright\arraybackslash}p{0.16\linewidth}
                >{\raggedright\arraybackslash}p{0.34\linewidth}
                >{\raggedright\arraybackslash}p{0.42\linewidth}@{}}
\caption{Representative atomic \benchname{} case examples by Word feature area.}
\label{tab:atomic-case-examples}\\
\toprule
\textbf{Word feature area} & \textbf{Source command} & \textbf{Instruction example} \\
\midrule
\endfirsthead
\toprule
\textbf{Word feature area} & \textbf{Source command} & \textbf{Instruction example} \\
\midrule
\endhead
\bottomrule
\endfoot
Text & \texttt{Bold} & Make the word 'Target' bold. \\
Paragraphs & \texttt{Align\allowbreak Center} & Center the paragraph containing 'Target'. \\
Lists & \texttt{Bullets\allowbreak Gallery\allowbreak Word} & Apply bullet list formatting to all paragraphs in the document using checkmark bullets. \\
Tables & \texttt{Table\allowbreak Insert\allowbreak Gallery} & Insert a 2x2 table. \\
Shape geometry & \texttt{Align\allowbreak Distribute\allowbreak Horizontally} & Distribute all three rectangles evenly horizontally with equal spacing between them. \\
Shape appearance & \texttt{Shape\allowbreak Fill\allowbreak Color\allowbreak Picker} & Set the shape's fill color to Blue (RGB: 0, 112, 192). \\
Shape text & \texttt{Text\allowbreak Align\allowbreak Gallery} & Align the text to the bottom inside the text box. \\
Pictures & \texttt{Picture\allowbreak Crop\allowbreak Aspect\allowbreak Ratio1\allowbreak To1} & Crop the image to a 1:1 (square) aspect ratio, centered on the image. \\
Graphics & \texttt{Icon\allowbreak Insert\allowbreak From\allowbreak File} & Insert a 150 x 150 pt star-shaped SVG icon filled with blue color (RGB: 68, 114, 196) into the document. \\
Charts & \texttt{Chart\allowbreak Insert} & Insert a clustered column chart into the document with categories 'Q1' and 'Q2', and one series 'Revenue' with values 100 and 150. \\
SmartArt & \texttt{Smart\allowbreak Art\allowbreak Insert} & Insert a SmartArt graphic using the 'Basic Block List' layout with nodes labeled 'Alpha', 'Beta', 'Charlie', 'Delta', and 'Echo'. \\
Equations & \texttt{Equation\allowbreak Fraction\allowbreak Gallery} & Add the fraction 'dy/dx' to the equation 'y ='. \\
Page Structure & \texttt{Page\allowbreak Break\allowbreak Insert\allowbreak Word} & Insert a page break after the word 'TARGET' to start a new page. \\
Annotations & \texttt{Insert\allowbreak New\allowbreak Comment} & Add a comment with text 'Test comment.' on the word 'TARGET'. \\
References & \texttt{Citation\allowbreak Insert} & At the beginning of the document body, insert an in-text citation using the existing source tagged 'Smith2020'. \\
\end{longtable}
}

\FloatBarrier

{
\scriptsize
\setlength{\tabcolsep}{3.5pt}
\renewcommand{\arraystretch}{1.13}
\begin{table}[!htbp]
\centering
\caption{Representative compositional \benchname{} case examples by composition class. Referenced atomic commands are derived from the case's linked \texttt{atomic\_intents}.}
\label{tab:compositional-case-examples}
\small
\begin{tabular}{@{}>{\raggedright\arraybackslash}p{0.17\linewidth}
                >{\raggedright\arraybackslash}p{0.34\linewidth}
                >{\raggedright\arraybackslash}p{0.41\linewidth}@{}}
\toprule
\textbf{Composition class} & \textbf{Referenced atomic commands} & \textbf{Instruction example} \\
\midrule
Additive & \makecell[tl]{\texttt{AlignCenter};\\\texttt{ParagraphSpacingBefore};\\\texttt{ParagraphSpacingAfter};\\\texttt{DropCapInsertGallery}} & Center the paragraph containing the word 'Target', set paragraph spacing to 0 points before and 0 points after for all paragraphs, and apply a dropped drop cap to the first paragraph. \\
Prerequisite & \makecell[tl]{\texttt{ChartInsert};\\\texttt{ChartDataLabel}} & Below the 'Quarterly Sales:' heading, insert a clustered column chart with categories 'Q1' and 'Q2', and one series 'Revenue' with values 100 and 150, then add data labels at the 'Outside End' position. \\
Scope-changing & \makecell[tl]{\texttt{TableRowsInsertBelowWord};\\\texttt{TableColumnsInsertRight};\\\texttt{TableSplitTable};\\\texttt{TableColumnsDistribute};\\\texttt{TableRowsDistribute}} & Insert a new row below the first row and a new column to the right of the first column, split the table at the inserted row, then distribute the columns in the upper table and distribute the rows in the lower table. \\
Layout-coupled & \makecell[tl]{\texttt{TextWrapGallery};\\\texttt{PicturePositionGallery}} & Apply square wrapping to the cross shape (text wraps around the cross's rectangular bounding box), then position it at the top-center of the page, using the Position preset (within default page margins). \\
\bottomrule
\end{tabular}
\end{table}
}

\FloatBarrier
\subsection{Complete Released-Command Catalogue}
\label{app:command_catalogue}
Table~\ref{tab:released-commands} enumerates the \nAtomicCommands{} released Atomic commands by Word feature area.

{\footnotesize
\setlength{\LTcapwidth}{\linewidth}
\begin{longtable}{@{}>{\raggedright\arraybackslash}p{72pt} >{\raggedright\arraybackslash}p{312pt}@{}}
\caption{Released atomic commands by Word feature area.}\label{tab:released-commands}\\
\toprule
\textbf{Word feature area} & \textbf{Released commands} \\
\midrule
\endfirsthead
\toprule
\textbf{Word feature area} & \textbf{Released commands} \\
\midrule
\endhead
\midrule\multicolumn{2}{r}{\emph{Continued on next page}}\\\endfoot
\bottomrule
\endlastfoot
Text & Bold, Change\allowbreak Case\allowbreak Gallery, Character\allowbreak Border, Character\allowbreak Shading, Clear\allowbreak Formatting, Font, Font\allowbreak Color\allowbreak Picker, Font\allowbreak Size, Font\allowbreak Size\allowbreak Decrease\allowbreak Word, Font\allowbreak Size\allowbreak Increase\allowbreak Word, Italic, Ligature\allowbreak Gallery\allowbreak Word, Shading\allowbreak Color\allowbreak Picker, Strikethrough, Subscript, Superscript, Symbol\allowbreak Insert\allowbreak Gallery, Text\allowbreak Fill\allowbreak Gradient\allowbreak Gallery, Text\allowbreak Highlight\allowbreak Color\allowbreak Picker, Text\allowbreak Outline\allowbreak Color\allowbreak Picker\allowbreak Alternate, Text\allowbreak Outline\allowbreak Dashes\allowbreak Gallery, Text\allowbreak Outline\allowbreak Weight\allowbreak Gallery, Text\allowbreak Reflection\allowbreak Gallery, Underline\allowbreak Color\allowbreak Picker, Underline\allowbreak Gallery \\ \addlinespace
Paragraphs & Align\allowbreak Center, Align\allowbreak Justify, Align\allowbreak Left, Align\allowbreak Right, Borders\allowbreak Selection\allowbreak Gallery, Drop\allowbreak Cap\allowbreak Insert\allowbreak Gallery, Indent\allowbreak Decrease\allowbreak Word, Indent\allowbreak Increase\allowbreak Word, Line\allowbreak Spacing\allowbreak Gallery, Paragraph\allowbreak Spacing\allowbreak After, Paragraph\allowbreak Spacing\allowbreak Before, Quick\allowbreak Styles\allowbreak Gallery, Table\allowbreak Of\allowbreak Contents\allowbreak Add\allowbreak Text\allowbreak Gallery \\ \addlinespace
Lists & Bullets\allowbreak Gallery\allowbreak Word, List\allowbreak Level\allowbreak Gallery, Multilevel\allowbreak List\allowbreak Gallery, Numbering\allowbreak Gallery\allowbreak Word \\ \addlinespace
Tables & Border\allowbreak Styles\allowbreak Gallery, Merge\allowbreak Cells, Quick\allowbreak Tables\allowbreak Insert\allowbreak Gallery, Sort\allowbreak Dialog\allowbreak Classic, Split\allowbreak Cells, Table\allowbreak Auto\allowbreak Fit\allowbreak Contents, Table\allowbreak Auto\allowbreak Fit\allowbreak Window, Table\allowbreak Cell\allowbreak Align\allowbreak Bottom\allowbreak Center, Table\allowbreak Cell\allowbreak Align\allowbreak Bottom\allowbreak Left, Table\allowbreak Cell\allowbreak Align\allowbreak Bottom\allowbreak Right, Table\allowbreak Cell\allowbreak Align\allowbreak Middle\allowbreak Center, Table\allowbreak Cell\allowbreak Align\allowbreak Middle\allowbreak Left, Table\allowbreak Cell\allowbreak Align\allowbreak Middle\allowbreak Right, Table\allowbreak Cell\allowbreak Align\allowbreak Top\allowbreak Center, Table\allowbreak Cell\allowbreak Align\allowbreak Top\allowbreak Left, Table\allowbreak Cell\allowbreak Align\allowbreak Top\allowbreak Right, Table\allowbreak Columns\allowbreak Delete, Table\allowbreak Columns\allowbreak Distribute, Table\allowbreak Columns\allowbreak Insert\allowbreak Left, Table\allowbreak Columns\allowbreak Insert\allowbreak Right, Table\allowbreak Delete, Table\allowbreak Insert\allowbreak Gallery, Table\allowbreak Rows\allowbreak Delete, Table\allowbreak Rows\allowbreak Distribute, Table\allowbreak Rows\allowbreak Insert\allowbreak Above\allowbreak Word, Table\allowbreak Rows\allowbreak Insert\allowbreak Below\allowbreak Word, Table\allowbreak Split\allowbreak Table, Table\allowbreak Style\allowbreak Clear, Table\allowbreak Styles\allowbreak Gallery\allowbreak Word \\ \addlinespace
Shape geometry & Align\allowbreak Distribute\allowbreak Horizontally, Align\allowbreak Distribute\allowbreak Vertically, Gallery\allowbreak All\allowbreak Shapes\allowbreak And\allowbreak Canvas, Object\allowbreak Bring\allowbreak Forward, Object\allowbreak Bring\allowbreak To\allowbreak Front, Object\allowbreak Rotate\allowbreak Gallery, Object\allowbreak Send\allowbreak Backward, Object\allowbreak Send\allowbreak To\allowbreak Back, Objects\allowbreak Align\allowbreak Bottom\allowbreak Smart, Objects\allowbreak Align\allowbreak Center\allowbreak Horizontal\allowbreak Smart, Objects\allowbreak Align\allowbreak Left\allowbreak Smart, Objects\allowbreak Align\allowbreak Middle\allowbreak Vertical\allowbreak Smart, Objects\allowbreak Align\allowbreak Right\allowbreak Smart, Objects\allowbreak Align\allowbreak Top\allowbreak Smart, Objects\allowbreak Group, Objects\allowbreak Ungroup, Picture\allowbreak Position\allowbreak Gallery, Shape\allowbreak Change\allowbreak Shape\allowbreak Gallery, Text\allowbreak Box\allowbreak Insert, Text\allowbreak Box\allowbreak Insert\allowbreak Gallery, Text\allowbreak Box\allowbreak Insert\allowbreak Vertical\allowbreak Word, Text\allowbreak Wrap\allowbreak Gallery \\ \addlinespace
Shape appearance & Bevel\allowbreak Shape\allowbreak Gallery, Glow\allowbreak Color\allowbreak Picker, Gradient\allowbreak Gallery, Object\allowbreak Effect\allowbreak Glow\allowbreak Gallery, Object\allowbreak Effect\allowbreak Shadow\allowbreak Gallery, Object\allowbreak Effect\allowbreak Soft\allowbreak Edges\allowbreak Gallery, Reflection\allowbreak Gallery, Shape\allowbreak Fill\allowbreak Color\allowbreak Picker, Shape\allowbreak Outline\allowbreak Color\allowbreak Picker, Shape\allowbreak Styles\allowbreak Gallery, \_3\allowbreak DRotation\allowbreak Gallery \\ \addlinespace
Shape text & Text\allowbreak Align\allowbreak Gallery, Text\allowbreak Direction, Text\allowbreak Direction\allowbreak Gallery \\ \addlinespace
Pictures & Horizontal\allowbreak Line\allowbreak Insert, Outline\allowbreak Color\allowbreak Picker, Picture\allowbreak Artistic\allowbreak Effects\allowbreak Gallery, Picture\allowbreak Color\allowbreak Tempertature\allowbreak Gallery, Picture\allowbreak Crop\allowbreak Aspect\allowbreak Ratio16\allowbreak To10, Picture\allowbreak Crop\allowbreak Aspect\allowbreak Ratio16\allowbreak To9, Picture\allowbreak Crop\allowbreak Aspect\allowbreak Ratio1\allowbreak To1, Picture\allowbreak Crop\allowbreak Aspect\allowbreak Ratio2\allowbreak To3, Picture\allowbreak Crop\allowbreak Aspect\allowbreak Ratio3\allowbreak To2, Picture\allowbreak Crop\allowbreak Aspect\allowbreak Ratio3\allowbreak To4, Picture\allowbreak Crop\allowbreak Aspect\allowbreak Ratio3\allowbreak To5, Picture\allowbreak Crop\allowbreak Aspect\allowbreak Ratio4\allowbreak To3, Picture\allowbreak Crop\allowbreak Aspect\allowbreak Ratio4\allowbreak To5, Picture\allowbreak Crop\allowbreak Aspect\allowbreak Ratio5\allowbreak To3, Picture\allowbreak Crop\allowbreak Aspect\allowbreak Ratio5\allowbreak To4, Picture\allowbreak Effects\allowbreak Glow\allowbreak Gallery, Picture\allowbreak Effects\allowbreak Shadow\allowbreak Gallery, Picture\allowbreak Effects\allowbreak Soft\allowbreak Edges\allowbreak Gallery, Picture\allowbreak Fill\allowbreak Crop, Picture\allowbreak Fit\allowbreak Crop, Picture\allowbreak Recolor\allowbreak Gallery, Picture\allowbreak Reflection\allowbreak Gallery, Picture\allowbreak Reset, Picture\allowbreak Reset\allowbreak And\allowbreak Size, Picture\allowbreak Rotation\allowbreak Gallery, Picture\allowbreak Saturation\allowbreak Gallery, Picture\allowbreak Sharpen\allowbreak Soften\allowbreak Gallery, \_3\allowbreak DBevel\allowbreak Picture\allowbreak Top\allowbreak Gallery \\ \addlinespace
Graphics & Graphics\allowbreak Color\allowbreak Picker, Graphics\allowbreak Outline\allowbreak Color\allowbreak Picker, Icon\allowbreak Insert\allowbreak From\allowbreak File \\ \addlinespace
Charts & Chart\allowbreak Colors\allowbreak Gallery, Chart\allowbreak Data\allowbreak Label, Chart\allowbreak Data\allowbreak Table, Chart\allowbreak Gridlines\allowbreak Gallery, Chart\allowbreak Insert, Chart\allowbreak Layout\allowbreak Gallery, Chart\allowbreak Legend, Chart\allowbreak Lines, Chart\allowbreak Reset\allowbreak To\allowbreak Match\allowbreak Style, Chart\allowbreak Switch\allowbreak Row\allowbreak Column, Chart\allowbreak Title, Chart\allowbreak Trendline, Chart\allowbreak Up\allowbreak Down\allowbreak Bars, Color\allowbreak Picker\allowbreak Chart\allowbreak Line\allowbreak Styles \\ \addlinespace
Smart\allowbreak Art & Arrow\allowbreak Style\allowbreak Gallery, Bevel\allowbreak Text\allowbreak Gallery, Outline\allowbreak Dashes\allowbreak Gallery, Outline\allowbreak Weight\allowbreak Gallery, Smart\allowbreak Art\allowbreak Add\allowbreak Assistant, Smart\allowbreak Art\allowbreak Add\allowbreak Shape\allowbreak Above, Smart\allowbreak Art\allowbreak Add\allowbreak Shape\allowbreak After, Smart\allowbreak Art\allowbreak Add\allowbreak Shape\allowbreak Before, Smart\allowbreak Art\allowbreak Add\allowbreak Shape\allowbreak Below, Smart\allowbreak Art\allowbreak Change\allowbreak Colors\allowbreak Gallery, Smart\allowbreak Art\allowbreak Demote, Smart\allowbreak Art\allowbreak Insert, Smart\allowbreak Art\allowbreak Larger\allowbreak Shape, Smart\allowbreak Art\allowbreak Layout\allowbreak Gallery, Smart\allowbreak Art\allowbreak Promote, Smart\allowbreak Art\allowbreak Reorder\allowbreak Down, Smart\allowbreak Art\allowbreak Reorder\allowbreak Up, Smart\allowbreak Art\allowbreak Reset\allowbreak Graphic, Smart\allowbreak Art\allowbreak Right\allowbreak To\allowbreak Left, Smart\allowbreak Art\allowbreak Smaller\allowbreak Shape, Smart\allowbreak Art\allowbreak Styles\allowbreak Gallery, Text\allowbreak Effect\allowbreak Transform\allowbreak Gallery, Text\allowbreak Effects3\allowbreak DRotation\allowbreak Gallery, Text\allowbreak Fill\allowbreak Color\allowbreak Picker \\ \addlinespace
Equations & Equation\allowbreak Accent\allowbreak Gallery, Equation\allowbreak Delimiter\allowbreak Gallery, Equation\allowbreak Fraction\allowbreak Gallery, Equation\allowbreak Function\allowbreak Gallery, Equation\allowbreak Insert\allowbreak New, Equation\allowbreak Integral\allowbreak Gallery, Equation\allowbreak Large\allowbreak Operator\allowbreak Gallery, Equation\allowbreak Limit\allowbreak Gallery, Equation\allowbreak Linear\allowbreak All, Equation\allowbreak Linear\allowbreak One, Equation\allowbreak Matrix\allowbreak Gallery, Equation\allowbreak Normal\allowbreak Text, Equation\allowbreak Operator\allowbreak Gallery, Equation\allowbreak Professional\allowbreak All, Equation\allowbreak Professional\allowbreak One, Equation\allowbreak Radical\allowbreak Gallery, Equation\allowbreak Script\allowbreak Gallery, Equation\allowbreak Symbols\allowbreak Insert\allowbreak Gallery, Insert\allowbreak Building\allowbreak Blocks\allowbreak Equations\allowbreak Gallery \\ \addlinespace
Page Structure & Blank\allowbreak Page\allowbreak Insert, Cover\allowbreak Page\allowbreak Insert\allowbreak Gallery, Doc\allowbreak Info\allowbreak Author, Doc\allowbreak Info\allowbreak Doc\allowbreak Title, Doc\allowbreak Info\allowbreak File\allowbreak Name, Footer\allowbreak Insert\allowbreak Gallery, Header\allowbreak Footer\allowbreak Remove\allowbreak Footer\allowbreak Word, Header\allowbreak Footer\allowbreak Remove\allowbreak Header\allowbreak Word, Header\allowbreak Insert\allowbreak Gallery, Page\allowbreak Break\allowbreak Insert\allowbreak Word, Page\allowbreak Nambers\allowbreak In\allowbreak Footer\allowbreak Insert\allowbreak Gallery, Page\allowbreak Nambers\allowbreak In\allowbreak Margins\allowbreak Insert\allowbreak Gallery, Page\allowbreak Numbers\allowbreak In\allowbreak Header\allowbreak Insert\allowbreak Gallery, Watermark\allowbreak Remove \\ \addlinespace
Annotations & Accept\allowbreak All\allowbreak Changes\allowbreak In\allowbreak Doc\allowbreak And\allowbreak Stop\allowbreak Tracking, Delete\allowbreak All\allowbreak Resolved\allowbreak Comments, Endnote\allowbreak Insert\allowbreak Word, Footnote\allowbreak Endnote\allowbreak Dialog, Footnote\allowbreak Insert, Insert\allowbreak New\allowbreak Comment, Reject\allowbreak All\allowbreak Changes\allowbreak In\allowbreak Doc\allowbreak And\allowbreak Stop\allowbreak Tracking, Review\allowbreak Delete\allowbreak All\allowbreak Comments\allowbreak In\allowbreak Document, Review\allowbreak Delete\allowbreak Comment, Review\allowbreak Track\allowbreak Changes \\ \addlinespace
References & Bibliography\allowbreak Gallery, Bibliography\allowbreak Insert, Bibliography\allowbreak Style, Caption\allowbreak Insert, Citation\allowbreak Insert, Hyperlink\allowbreak Edit, Hyperlink\allowbreak Insert, Hyperlink\allowbreak Remove, Index\allowbreak Update, Page\allowbreak Number\allowbreak Field\allowbreak Insert\allowbreak Gallery, Table\allowbreak Of\allowbreak Contents\allowbreak Gallery, Table\allowbreak Of\allowbreak Contents\allowbreak Remove \\
\end{longtable}
}

\FloatBarrier

%% file: assets/tables/object_property_coverage.tex
{\footnotesize
\setlength{\LTcapwidth}{\linewidth}
\begin{longtable}{@{}l >{\raggedright\arraybackslash}p{0.20\linewidth} >{\raggedright\arraybackslash}p{0.56\linewidth}@{}}
\caption{Object--property coverage of the Word-native oracle, organized by Word feature area.}\label{tab:capability-hierarchy}\\
\toprule
\textbf{Feature area} & \textbf{Runtime object} & \textbf{Object properties} \\
\midrule
\endfirsthead
\toprule
\textbf{Feature area} & \textbf{Runtime object} & \textbf{Object properties} \\
\midrule
\endhead
\bottomrule
\endfoot
Text & Border & Color, LineStyle, LineWidth \\
 & ColorFormat & RGB \\
 & FillFormat & GradientAngle, Type \\
 & Font & Bold, Color, Italic, Ligatures, Name, Size, StrikeThrough, Subscript, Superscript, Underline, UnderlineColor \\
 & LineFormat & DashStyle, Weight \\
 & Range & HighlightColorIndex, Text \\
 & ReflectionFormat & Type \\
 & Shading & BackgroundPatternColor, Texture \\
\addlinespace
Paragraphs & Border & LineStyle \\
 & DropCap & LinesToDrop, Position \\
 & Paragraph & OutlineLevel \\
 & ParagraphFormat & Alignment, LeftIndent, LineSpacing, LineSpacingRule, SpaceAfter, SpaceBefore \\
 & Style & NameLocal \\
\addlinespace
Lists & ListFormat & ListLevelNumber, ListString \\
\addlinespace
Tables & Border & LineStyle \\
 & Cell & VerticalAlignment \\
 & Column & Width \\
 & Columns & Count \\
 & Paragraph & Alignment \\
 & Range & Text \\
 & Row & Height \\
 & Rows & Count \\
 & Style & NameLocal \\
 & Table & AutoFitBehavior \\
 & Tables & Count \\
\addlinespace
Shape geometry & GroupShapes & Count \\
 & Shape & AutoShapeType, Height, HorizontalFlip, Left, RelativeHorizontalPosition, RelativeVerticalPosition, Rotation, Top, Type, VerticalFlip, Width, ZOrderPosition \\
 & Shapes & Count \\
 & WrapFormat & Type \\
\addlinespace
Shape appearance & ColorFormat & RGB \\
 & FillFormat & GradientStops, Type \\
 & GlowFormat & Radius \\
 & ReflectionFormat & Type \\
 & ShadowFormat & Type \\
 & Shape & ShapeStyle \\
 & SoftEdgeFormat & Type \\
 & ThreeDFormat & BevelTopDepth, BevelTopType, PresetCamera, RotationX, RotationY \\
\addlinespace
Shape text & Paragraph & Alignment \\
 & Range & Orientation, Text \\
 & TextFrame & Orientation \\
 & TextFrame2 & Orientation \\
\addlinespace
Pictures & ColorFormat & RGB \\
 & GlowFormat & Radius \\
 & InlineShape & Height, Type, Width \\
 & InlineShapes & Count \\
 & PictureFormat & Brightness, ColorType, Contrast, CropBottom, CropLeft, CropRight, CropTop \\
 & ReflectionFormat & Type \\
 & ShadowFormat & Type \\
 & SoftEdgeFormat & Type \\
 & ThreeDFormat & BevelTopDepth, BevelTopType, PresetCamera, RotationX, RotationY \\
\addlinespace
Graphics & ColorFormat & RGB \\
 & InlineShape & Height, Width \\
 & InlineShapes & Count \\
\addlinespace
Charts & Axis & MajorGridlines, MinorGridlines \\
 & Chart & ChartColor, ChartGroups, ChartStyle, ChartType, HasDataTable, HasLegend, HasTitle \\
 & ChartTitle & Text \\
 & ColorFormat & RGB \\
 & DataLabels & Position \\
 & DataTable & ShowLegendKey \\
 & Legend & Position \\
 & Series & Format, HasDataLabels, Name, Values \\
 & SeriesCollection & Count \\
 & Trendline & Type \\
\addlinespace
SmartArt & ColorFormat & RGB \\
 & LineFormat & BeginArrowheadStyle, DashStyle, EndArrowheadStyle, Weight \\
 & Shape & Height, Width \\
 & SmartArt & Reverse \\
 & SmartArtColor & Name \\
 & SmartArtLayout & Name \\
 & SmartArtNode & Level \\
 & SmartArtNodes & Count \\
 & SmartArtQuickStyle & Name \\
 & TextFrame2 & WarpFormat \\
 & TextRange2 & Text \\
 & ThreeDFormat & BevelTopType, PresetCamera, RotationX \\
\addlinespace
Equations & OMaths & Count \\
 & Range & Text, WordOpenXML \\
\addlinespace
Page Structure & ContentControl & Tag, Title \\
 & ContentControls & Count \\
 & Document & ComputeStatistics, Content \\
 & ParagraphFormat & Alignment \\
 & Range & Text \\
 & Shapes & Count \\
 & Style & NameLocal \\
\addlinespace
Annotations & Comment & Author, Done, Index \\
 & Comments & Count \\
 & Document & TrackRevisions \\
 & Endnotes & Count, NumberStyle \\
 & Footnotes & Count, NumberingRule, NumberStyle \\
 & Range & Start, Text \\
 & Revisions & Count \\
\addlinespace
References & Bibliography & BibliographyStyle \\
 & ContentControl & Title \\
 & ContentControls & Count \\
 & Hyperlink & Address, TextToDisplay \\
 & Hyperlinks & Count \\
 & Indexes & Count \\
 & Range & Text \\
 & Sources & Count \\
 & TableOfContents & IncludePageNumbers \\
 & TablesOfContents & Count \\
\addlinespace
\end{longtable}
}

%% file: assets/tables/human_oracle_audit.tex
\begin{table}[!htbp]
\centering
\caption{Evaluator judgments against the original human majority on 258 valid documents (167 Pass, 91 Fail). Pass is the positive class; all values are percentages.}
\label{tab:human-oracle-audit}
\small
\setlength{\tabcolsep}{6pt}
\begin{tabular}{@{}lrrr@{}}
\toprule
\textbf{Evaluator} & \textbf{Precision (\%)} $\uparrow$ & \textbf{FPR (\%)} $\downarrow$ & \textbf{FNR (\%)} $\downarrow$ \\
\midrule
Word Object & 96.5 & 5.5 & 16.8 \\
Word Revision & 97.1 & 2.2 & 59.9 \\
SSIM $\geq 0.99$ & 77.3 & 45.1 & 16.2 \\
SSIM $\geq 0.999$ & 94.5 & 7.7 & 27.5 \\
agent judge & 93.0 & 12.1 & 12.0 \\
\bottomrule
\end{tabular}
\end{table}

%% file: assets/tables/agent_versions.tex
\begin{table}[h]
\centering
\caption{Agent packages and model identifiers. -- denotes an unavailable version.}
\label{tab:agent-versions}
\small
\begin{tabular}{@{}llll@{}}
\toprule
\textbf{Agent system} & \textbf{Package} & \textbf{Version} & \textbf{Model identifier} \\
\midrule
Claude Code & \texttt{@anthropic-ai/\allowbreak{}claude-code} & -- & \texttt{Opus 5} \\
Codex & \texttt{@openai/\allowbreak{}codex} & 0.128.0 & \texttt{gpt-5.6-sol} \\
Kimi Code & \texttt{kimi-cli} & 1.41.0 & \texttt{Kimi-K3} \\
OpenCode & \texttt{opencode} & 1.18.3 & \texttt{gemini-3.8-flash} \\
ZCode & \texttt{zcode-agent-cli} & 0.15.2 & \texttt{GLM-5.2} \\
OpenCode & \texttt{opencode} & -- & \texttt{Deep\allowbreak{}Seek-V4-Pro} \\
Qwen Code & \texttt{@qwen-code/\allowbreak{}qwen-code} & 0.22.3 & \texttt{Qwen/\allowbreak{}Qwen3.8-27B} \\
Mini Agent & \texttt{Mini-Agent} & -- & \texttt{Mini\allowbreak{}Max M3} \\
\bottomrule
\end{tabular}
\end{table}

%% file: assets/tables/human_execution.tex
\begin{table}[!htbp]
\centering
\caption{Human execution on the same 642 tasks per participant. Case denominators are 546 Atomic, 96 Compositional and 642 Overall; Macro uses 231, 27 and 258 fixed-variant groups. The main-table Human row averages the three participant-specific rates.}
\label{tab:human-execution}
\small
\setlength{\tabcolsep}{3pt}
\begin{tabular}{@{}l@{\hspace{12pt}}
    *{2}{>{\raggedleft\arraybackslash}p{31pt}}@{\hspace{9pt}}
    *{2}{>{\raggedleft\arraybackslash}p{31pt}}@{\hspace{9pt}}
    *{2}{>{\raggedleft\arraybackslash}p{31pt}}@{}}
\toprule
\textbf{Participant} & \multicolumn{2}{c}{\textbf{Atomic}} & \multicolumn{2}{c}{\textbf{Compositional}} & \multicolumn{2}{c}{\textbf{Overall}} \\
\cmidrule(lr){2-3}\cmidrule(lr){4-5}\cmidrule(l){6-7}
 & Case & Macro & Case & Macro & Case & Macro \\
\midrule
Human A01 & 83.0 & 85.4 & 71.9 & 69.6 & 81.3 & 83.7 \\
Human A02 & 90.1 & 90.9 & 78.1 & 76.8 & 88.3 & 89.4 \\
Human A03 & 82.4 & 83.9 & 58.3 & 58.8 & 78.8 & 81.3 \\
\bottomrule
\end{tabular}
\end{table}

%% file: assets/tables/composition_outcome_decomposition.tex
\begin{table}[h]
\centering
\caption{Pooled intent outcomes by composition class. Rates use observed pairs across the eight systems ($n_{\text{intent}}$). Operation-limited and coincidence are the shares of $\bar{A}\bar{C}$ and $\bar{A}$C pairs, and Transfer fail is the share of A$\bar{C}$ among Atomic-pass pairs.}
\label{tab:intent-outcomes}
\small
\setlength{\tabcolsep}{3pt}
\begin{tabular}{@{}l rrrrr r@{}}
\toprule
\textbf{Class} & $\Pr(A_i)$ & Transfer & Transfer fail & Operation-limited & Coincidence & $n_{\text{intent}}$ \\
\midrule
Additive & 79.7 & 91.2 & 8.8 & 9.7 & 10.7 & 1170 \\
Prerequisite & 70.8 & 93.5 & 6.5 & 8.2 & 21.0 & 219 \\
Scope-changing & 91.9 & 95.2 & 4.8 & 1.5 & 6.6 & 136 \\
Layout-coupled & 90.9 & 71.9 & 28.1 & 1.7 & 7.4 & 176 \\
\bottomrule
\end{tabular}
\end{table}

%% file: assets/tables/composition_case_conditioning.tex
\begin{table}[h]
\centering
\caption{Case-level and intent-level rates by system and composition class. Case-level rates use all assigned cases (46, 25, 12, and 13 per class) and count missing outputs as failures. $P(A)$ is the share of cases whose linked Atomic variants all pass, and $n_A$ counts them. $P(C\mid A)$ is omitted (--) when fewer than 5 such cases exist. Intent-level rates use observed pairs, the pass gap is $\Pr(A_i)-\Pr(C_i)$, and $n_{C\mid A}$ counts Atomic-pass intents.}
\label{tab:case-conditioning-denominators}
\footnotesize
\setlength{\tabcolsep}{3.3pt}
\begin{tabular}{@{}l l rrrr rrrrr@{}}
\toprule
 & & \multicolumn{4}{c}{\textbf{Case level}} & \multicolumn{5}{c}{\textbf{Intent level}} \\
\cmidrule(lr){3-6}\cmidrule(l){7-11}
\textbf{System} & \textbf{Class} & $P(A)$ & $P(C)$ & $P(C \mid A)$ & $n_A$ & $\Pr(A_i)$ & $\Pr(C_i)$ & Pass gap & $\Pr(C_i \mid A_i)$ & $n_{C\mid A}$ \\
\midrule
Opus 5 & Additive & 80.4 & 82.6 & 89.2 & 37 & 97.4 & 93.8 & 3.6 & 93.7 & 190 \\
 & Prerequisite & 80.0 & 80.0 & 85.0 & 20 & 95.5 & 95.5 & 0.0 & 97.6 & 42 \\
 & Scope-changing & 58.3 & 100.0 & 100.0 & 7 & 95.8 & 100.0 & -4.2 & 100.0 & 23 \\
 & Layout-coupled & 100.0 & 84.6 & 84.6 & 13 & 100.0 & 93.8 & 6.2 & 93.8 & 32 \\
\midrule
GPT-5.6 & Additive & 37.0 & 56.5 & 58.8 & 17 & 73.2 & 90.9 & -17.7 & 91.7 & 120 \\
 & Prerequisite & 44.0 & 52.0 & 54.5 & 11 & 72.7 & 90.9 & -18.2 & 95.8 & 24 \\
 & Scope-changing & 58.3 & 83.3 & 100.0 & 7 & 90.0 & 100.0 & -10.0 & 100.0 & 18 \\
 & Layout-coupled & 69.2 & 76.9 & 66.7 & 9 & 85.7 & 92.9 & -7.1 & 91.7 & 24 \\
\midrule
K3 & Additive & 41.3 & 47.8 & 89.5 & 19 & 85.6 & 82.7 & 2.9 & 89.9 & 119 \\
 & Prerequisite & 28.0 & 48.0 & 42.9 & 7 & 60.0 & 88.6 & -28.6 & 95.2 & 21 \\
 & Scope-changing & 75.0 & 75.0 & 66.7 & 9 & 95.2 & 95.2 & 0.0 & 95.0 & 20 \\
 & Layout-coupled & 38.5 & 61.5 & 80.0 & 5 & 94.4 & 94.4 & 0.0 & 94.1 & 17 \\
\midrule
Gemini 3.8 Flash & Additive & 28.3 & 45.7 & 76.9 & 13 & 96.2 & 100.0 & -3.8 & 100.0 & 76 \\
 & Prerequisite & 40.0 & 48.0 & 60.0 & 10 & 100.0 & 100.0 & 0.0 & 100.0 & 18 \\
 & Scope-changing & 25.0 & 58.3 & -- & 3 & 100.0 & 100.0 & 0.0 & 100.0 & 6 \\
 & Layout-coupled & 15.4 & 53.8 & -- & 2 & 100.0 & 91.7 & 8.3 & 91.7 & 12 \\
\midrule
GLM-5.2 & Additive & 41.3 & 43.5 & 73.7 & 19 & 82.8 & 85.9 & -3.1 & 93.4 & 106 \\
 & Prerequisite & 16.0 & 32.0 & -- & 4 & 45.5 & 86.4 & -40.9 & 90.0 & 10 \\
 & Scope-changing & 41.7 & 58.3 & 100.0 & 5 & 88.9 & 94.4 & -5.6 & 100.0 & 16 \\
 & Layout-coupled & 61.5 & 23.1 & 25.0 & 8 & 90.5 & 47.6 & 42.9 & 42.1 & 19 \\
\midrule
DeepSeek V4 Pro & Additive & 39.1 & 39.1 & 61.1 & 18 & 69.1 & 79.2 & -10.1 & 87.0 & 123 \\
 & Prerequisite & 28.0 & 36.0 & 57.1 & 7 & 64.3 & 82.1 & -17.9 & 88.9 & 18 \\
 & Scope-changing & 83.3 & 66.7 & 70.0 & 10 & 90.5 & 85.7 & 4.8 & 89.5 & 19 \\
 & Layout-coupled & 53.8 & 30.8 & 28.6 & 7 & 84.0 & 60.0 & 24.0 & 52.4 & 21 \\
\midrule
Qwen3.8-27B & Additive & 37.0 & 30.4 & 64.7 & 17 & 78.5 & 70.2 & 8.3 & 83.2 & 95 \\
 & Prerequisite & 40.0 & 24.0 & 30.0 & 10 & 63.6 & 77.3 & -13.6 & 85.7 & 14 \\
 & Scope-changing & 33.3 & 66.7 & -- & 4 & 86.7 & 93.3 & -6.7 & 92.3 & 13 \\
 & Layout-coupled & 46.2 & 15.4 & 16.7 & 6 & 91.7 & 37.5 & 54.2 & 36.4 & 22 \\
\midrule
MiniMax M3 & Additive & 26.1 & 34.8 & 83.3 & 12 & 62.0 & 68.1 & -6.0 & 91.3 & 103 \\
 & Prerequisite & 8.0 & 16.0 & -- & 2 & 47.1 & 64.7 & -17.6 & 75.0 & 8 \\
 & Scope-changing & 58.3 & 41.7 & 42.9 & 7 & 90.9 & 81.8 & 9.1 & 80.0 & 10 \\
 & Layout-coupled & 38.5 & 23.1 & 40.0 & 5 & 81.2 & 62.5 & 18.8 & 69.2 & 13 \\
\bottomrule
\end{tabular}
\end{table}

%% file: assets/tables/composition_transfer_significance.tex
\begin{table}[t]
\centering
\caption{Closed- versus open-weight transfer by composition class. $\Delta$ is closed minus open in percentage points with its 95\% paired bootstrap CI, $n$ counts Atomic-pass intents, and clusters counts the cases observed in both groups.}
\label{tab:burden-tiers}
\small
\setlength{\tabcolsep}{3pt}
\begin{tabular}{@{}l rrrr rr r@{}}
\toprule
\textbf{Class} & closed (\%) & open (\%) & $\Delta$ (pp) & 95\% CI & $n_{\text{closed}}$ & $n_{\text{open}}$ & clusters \\
\midrule
Additive & 94.3 & 88.9 & 5.3 & [1.2, 10.2] & 384 & 542 & 44 \\
Prerequisite & 97.0 & 88.7 & 8.2 & [0.0, 18.1] & 66 & 71 & 19 \\
Scope-changing & 100.0 & 92.3 & 7.7 & [0.0, 18.4] & 47 & 78 & 12 \\
Layout-coupled & 92.2 & 56.5 & 35.7 & [19.5, 53.7] & 64 & 92 & 12 \\
\bottomrule
\end{tabular}
\end{table}

%% file: assets/tables/composition_transfer_by_operation.tex
\begin{table}[!htbp]
\centering
\caption{Twelve lowest-transfer operations with at least 15 observed pairs, pooled across the eight systems. $n_{\text{intent}}$ counts observed pairs.}
\label{tab:burden-operation}
\small
\setlength{\tabcolsep}{4pt}
\begin{tabular}{@{}l rrrr r@{}}
\toprule
\textbf{Operation} & $n_{\text{intent}}$ & $\Pr(A_i)$ & $\Pr(C_i)$ & $\Pr(C_i \mid A_i)$ & $n_{C\mid A}$ \\
\midrule
\texttt{shadow} & 19 & 89.5 & 26.3 & 29.4 & 17 \\
\texttt{pic\_\allowbreak{}shadow} & 19 & 84.2 & 26.3 & 31.2 & 16 \\
\texttt{reflection} & 17 & 52.9 & 35.3 & 44.4 & 9 \\
\texttt{rotate\_\allowbreak{}angle} & 46 & 78.3 & 52.2 & 47.2 & 36 \\
\texttt{flip\_\allowbreak{}direction} & 17 & 100.0 & 58.8 & 58.8 & 17 \\
\texttt{align\_\allowbreak{}command} & 73 & 100.0 & 72.6 & 72.6 & 73 \\
\texttt{list\_\allowbreak{}type} & 20 & 80.0 & 65.0 & 75.0 & 16 \\
\texttt{insert\_\allowbreak{}icon} & 17 & 76.5 & 88.2 & 92.3 & 13 \\
\texttt{distribute\_\allowbreak{}command} & 47 & 100.0 & 93.6 & 93.6 & 47 \\
\texttt{position\_\allowbreak{}preset} & 22 & 95.5 & 90.9 & 95.2 & 21 \\
\texttt{legend} & 30 & 90.0 & 96.7 & 96.3 & 27 \\
\texttt{alignment} & 46 & 100.0 & 100.0 & 100.0 & 46 \\
\bottomrule
\end{tabular}
\end{table}

%% file: assets/tables/composition_transfer_by_subsystem.tex
\begin{table}[!htbp]
\centering
\caption{Intent transfer by Word feature area, pooled across the eight systems. $n_{\text{intent}}$ counts observed pairs.}
\label{tab:burden-subsystem}
\small
\setlength{\tabcolsep}{4pt}
\begin{tabular}{@{}l rrrr r@{}}
\toprule
\textbf{Word feature area} & $n_{\text{intent}}$ & $\Pr(A_i)$ & $\Pr(C_i)$ & $\Pr(C_i \mid A_i)$ & $n_{C\mid A}$ \\
\midrule
Annotations & 51 & 84.3 & 98.0 & 97.7 & 43 \\
Charts & 171 & 88.9 & 95.3 & 97.4 & 152 \\
Equations & 75 & 61.3 & 92.0 & 93.5 & 46 \\
Graphics & 17 & 76.5 & 88.2 & 92.3 & 13 \\
Lists & 48 & 87.5 & 77.1 & 83.3 & 42 \\
Page Structure & 49 & 73.5 & 69.4 & 91.7 & 36 \\
Paragraphs & 225 & 89.3 & 96.9 & 99.5 & 201 \\
Pictures & 128 & 59.4 & 46.9 & 61.8 & 76 \\
References & 14 & 71.4 & 78.6 & 90.0 & 10 \\
Shape appearance & 130 & 80.8 & 72.3 & 85.7 & 105 \\
Shape geometry & 286 & 89.2 & 79.0 & 79.6 & 255 \\
Shape text & 13 & 84.6 & 84.6 & 81.8 & 11 \\
SmartArt & 128 & 74.2 & 71.9 & 88.4 & 95 \\
Tables & 133 & 85.7 & 93.2 & 94.7 & 114 \\
Text & 233 & 74.2 & 93.6 & 96.0 & 173 \\
\bottomrule
\end{tabular}
\end{table}

%% file: assets/tables/composition_trivial_pass_examples.tex
\begin{table}[!htbp]
\centering
\caption{End-state coincidence examples by operation. Each row is a case with an Atomic Fail, embedded Pass, and whole-case Pass.}
\label{tab:burden-trivial}
\small
\setlength{\tabcolsep}{3pt}
\begin{tabular}{@{}>{\raggedright\arraybackslash}p{0.19\linewidth} >{\raggedright\arraybackslash}p{0.13\linewidth} >{\raggedright\arraybackslash}p{0.12\linewidth} >{\raggedright\arraybackslash}p{0.29\linewidth} >{\raggedright\arraybackslash}p{0.20\linewidth}@{}}
\toprule
\textbf{Operation} & \textbf{System} & \textbf{Class} & \textbf{Compositional case} & \textbf{Atomic case} \\
\midrule
\texttt{insert\_\allowbreak{}equation} & GPT-5.6 & Prerequisite & \texttt{Prerequisite\allowbreak{}Equation/\allowbreak{}variant\_\allowbreak{}fraction} & \texttt{Equation\allowbreak{}Insert\allowbreak{}New/\allowbreak{}apply} \\
\texttt{drop\_\allowbreak{}cap} & GLM-5.2 & Additive & \texttt{Additive\allowbreak{}Paragraph\allowbreak{}Composition/\allowbreak{}variant\_\allowbreak{}center\_\allowbreak{}dropcap} & \texttt{Drop\allowbreak{}Cap\allowbreak{}Insert\allowbreak{}Gallery/\allowbreak{}dropped} \\
\texttt{outline\_\allowbreak{}weight} & MiniMax M3 & Additive & \texttt{Additive\allowbreak{}Font\allowbreak{}Composition/\allowbreak{}variant\_\allowbreak{}mixed\_\allowbreak{}effects} & \texttt{Text\allowbreak{}Outline\allowbreak{}Weight\allowbreak{}Gallery/\allowbreak{}apply\_\allowbreak{}1pt} \\
\texttt{outline\_\allowbreak{}color} & K3 & Additive & \texttt{Additive\allowbreak{}Font\allowbreak{}Composition/\allowbreak{}variant\_\allowbreak{}off\_\allowbreak{}heavy} & \texttt{Text\allowbreak{}Outline\allowbreak{}Color\allowbreak{}Picker\allowbreak{}Alternate/\allowbreak{}apply\_\allowbreak{}Blue} \\
\texttt{rotate\_\allowbreak{}angle} & GPT-5.6 & Additive & \texttt{Additive\allowbreak{}Layout\allowbreak{}Composition/\allowbreak{}variant\_\allowbreak{}rotate\_\allowbreak{}cw\_\allowbreak{}distribute\_\allowbreak{}h} & \texttt{Object\allowbreak{}Rotate\allowbreak{}Gallery/\allowbreak{}apply\_\allowbreak{}rotate\_\allowbreak{}right\_\allowbreak{}90} \\
\texttt{table\_\allowbreak{}style} & Qwen3.8-27B & Additive & \texttt{Additive\allowbreak{}Table\allowbreak{}Composition/\allowbreak{}variant\_\allowbreak{}style\_\allowbreak{}textdir\_\allowbreak{}border} & \texttt{Table\allowbreak{}Styles\allowbreak{}Gallery\allowbreak{}Word/\allowbreak{}Grid\_\allowbreak{}Table\_\allowbreak{}4\_\allowbreak{}\_\allowbreak{}\_\allowbreak{}Accent\_\allowbreak{}1} \\
\texttt{ungroup\_\allowbreak{}command} & DeepSeek V4 Pro & Layout-coupled & \texttt{Layout\allowbreak{}Coupled\allowbreak{}Group\allowbreak{}Lifecycle/\allowbreak{}variant\_\allowbreak{}flip\_\allowbreak{}horizontal} & \texttt{Objects\allowbreak{}Ungroup/\allowbreak{}apply} \\
\bottomrule
\end{tabular}
\end{table}